\documentclass[journal]{IEEEtran} %usletter, 10 pt

\usepackage{eurosym}
\usepackage{upgreek}
\usepackage{cite}

\ifCLASSINFOpdf
  \usepackage[pdftex]{graphicx}
  \usepackage{import}

\usepackage[english]{babel} 
\usepackage[utf8]{inputenc}
\usepackage[]{caption}
\usepackage{subcaption}
\usepackage{placeins}
\usepackage{color}
\usepackage{mathtools}
\usepackage{placeins}
\usepackage{relsize}
\usepackage{amssymb}
\usepackage{units}
\usepackage{amsmath}
\usepackage{float} 		
\renewcommand{\figurename}{Fig.}
\addto\captionsenglish{\renewcommand{\figurename}{Fig.}}
\usepackage{nicefrac}
\usepackage{tikz}
\usepackage[percent]{overpic}
\usepackage[hidelinks]{hyperref}

\definecolor{mygray}{RGB}{70, 70, 70}
\definecolor{mygray2}{RGB}{40, 40, 40}
\definecolor{myred}{RGB}{255, 53, 53}
\definecolor{myblue}{RGB}{0, 70, 254}
\definecolor{myyellow}{RGB}{247, 219, 0}

\usepackage{tikz}
\usepackage[color=black,opacity=1,angle=0,scale=1]{background}
\backgroundsetup{
contents={
\begin{tikzpicture}
\node at (current page.center) [align=center] {\textcopyright 2021 IEEE. Personal use of this material is permitted. 
\\ Permission from IEEE must be obtained for all other uses, in any current or future media, including reprinting/republishing this material for advertising or promotional purposes, 
\\ creating new collective works, for resale or redistribution to servers or lists, or reuse of any copyrighted component of this work in other works.
\\ DOI: 10.1109/TIV.2021.3091188};
\end{tikzpicture}},
placement=bottom,
scale=0.6,
vshift=20
}

\begin{document}
\title{Online and Predictive Warning System \\\vspace{-0.0001cm} for Forced Lane Changes using Risk Maps} 
\author{Tim Puphal, Benedict Flade, Malte Probst, Volker Willert, J\"urgen Adamy and Julian Eggert \vspace{-0.25cm}
\thanks{T. Puphal, B. Flade, M. Probst and J. Eggert are with the Honda Research Institute (HRI) Europe, Carl-Legien-Str. 30, 63073 Offenbach, Germany. Email: tim.puphal@honda-ri.de 

V. Willert is with the Faculty of Electrical Engineering (FHWS), Konrad-Geiger-Str. 2, 97421 Schweinfurt, Germany. 

J. Adamy is with the Control Methods and Robotics Lab (TU Darmstadt), Landgraf-Georg Str. 4, 64283 Darmstadt, Germany.
}} 
\maketitle

\begin{abstract} 
The survival analysis of driving trajectories allows for holistic evaluations of car-related risks caused by collisions or curvy roads. This analysis has advantages over common Time-To-X indicators, such as its predictive and probabilistic nature. However, so far, the theoretical risks have not been demonstrated in real-world environments. In this paper, we therefore present Risk Maps (RM) for online warning support in situations with forced lane changes, due to the end of roads.

For this purpose, we first unify sensor data in a Relational Local Dynamic Map (R-LDM). RM is afterwards able to be run in real-time and efficiently probes a range of situations in order to determine risk-minimizing behaviors. Hereby, we focus on the improvement of uncertainty-awareness and transparency of the system. Risk, utility and comfort costs are included in a single formula and are intuitively visualized to the driver. 

In the conducted experiments, a low-cost sensor setup with a GNSS receiver for localization and multiple cameras for object detection are leveraged. The final system is successfully applied on two-lane roads and recommends lane change advices, which are separated in gap and no-gap indications. These results are promising and present an important step towards interpretable safety. 
\end{abstract}

% For peerreview papers, this IEEEtran command inserts a page break and
% creates the second title. It will be ignored for other modes.
\IEEEpeerreviewmaketitle


\section{Introduction}
\label{sec:intro}
\vspace{0.05cm}

\IEEEPARstart{A}{dvanced} Driver Assistance Systems (ADAS) aim for supporting the human driver in its driving task \cite{bengler2014} and serve to avoid potential vehicle accidents. Prominent examples of ADAS are, e.g., the adaptive cruise control \cite{winner1996} and various lane keeping aids \cite{gayko2004}. They successfully help drivers on highway scenarios. Still, most of the current ADAS are not incorporating explicit driving predictions and uncertainty considerations in their planning.  Especially, in real traffic situations with high complexity, the technical requirements for ADAS robustness are high. Such situations include, e.g., dynamic lane changing or tailgaiting behaviors. Here, ADAS that allow to visualize the situation understanding modules in terms of driving risks are beneficial. A technical system must show reasons behind its decisions, which will eventually also increase the trust of the user. 

\begin{figure}[t]
  \centering
  \vspace{0.63cm}
  \resizebox{0.87\linewidth}{!}{%% Creator: Inkscape inkscape 0.92.4, www.inkscape.org
%% PDF/EPS/PS + LaTeX output extension by Johan Engelen, 2010
%% Accompanies image file 'highway_entrance_new.pdf' (pdf, eps, ps)
%%
%% To include the image in your LaTeX document, write
%%   \input{<filename>.pdf_tex}
%%  instead of
%%   \includegraphics{<filename>.pdf}
%% To scale the image, write
%%   \def\svgwidth{<desired width>}
%%   \input{<filename>.pdf_tex}
%%  instead of
%%   \includegraphics[width=<desired width>]{<filename>.pdf}
%%
%% Images with a different path to the parent latex file can
%% be accessed with the `import' package (which may need to be
%% installed) using
%%   \usepackage{import}
%% in the preamble, and then including the image with
%%   \import{<path to file>}{<filename>.pdf_tex}
%% Alternatively, one can specify
%%   \graphicspath{{<path to file>/}}
%% 
%% For more information, please see info/svg-inkscape on CTAN:
%%   http://tug.ctan.org/tex-archive/info/svg-inkscape
%%
\begingroup%
  \makeatletter%
  \providecommand\color[2][]{%
    \errmessage{(Inkscape) Color is used for the text in Inkscape, but the package 'color.sty' is not loaded}%
    \renewcommand\color[2][]{}%
  }%
  \providecommand\transparent[1]{%
    \errmessage{(Inkscape) Transparency is used (non-zero) for the text in Inkscape, but the package 'transparent.sty' is not loaded}%
    \renewcommand\transparent[1]{}%
  }%
  \providecommand\rotatebox[2]{#2}%
  \newcommand*\fsize{\dimexpr\f@size pt\relax}%
  \newcommand*\lineheight[1]{\fontsize{\fsize}{#1\fsize}\selectfont}%
  \ifx\svgwidth\undefined%
    \setlength{\unitlength}{260.21773499bp}%
    \ifx\svgscale\undefined%
      \relax%
    \else%
      \setlength{\unitlength}{\unitlength * \real{\svgscale}}%
    \fi%
  \else%
    \setlength{\unitlength}{\svgwidth}%
  \fi%
  \global\let\svgwidth\undefined%
  \global\let\svgscale\undefined%
  \makeatother%
  \begin{picture}(1,0.19034839)%
    \lineheight{1}%
    \setlength\tabcolsep{0pt}%
    \put(0,0){\includegraphics[width=\unitlength,page=1]{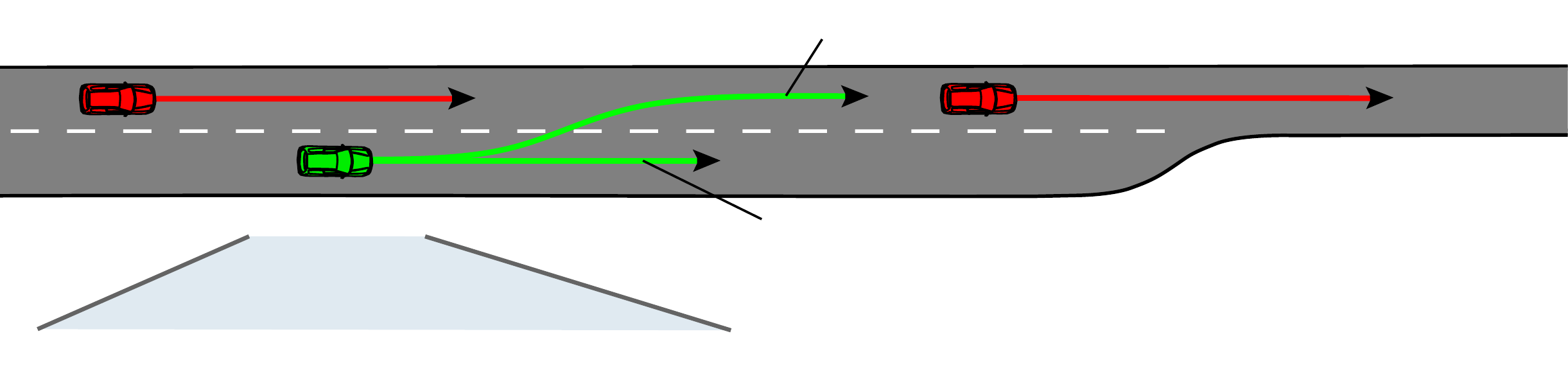}}%
    \put(0.43604817,0.06901373){\color[rgb]{0,0,0}\makebox(0,0)[lt]{\lineheight{1.25}\smash{\begin{tabular}[t]{l}stay on lane and brake\end{tabular}}}}%
	\put(0.14404817,0.05251373){\color[rgb]{0,0,0}\makebox(0,0)[lt]{\lineheight{1.25}\smash{\begin{tabular}[t]{l}{\fontsize{9.45}{11}\selectfont \textcolor{black}{\textit{Risk Maps}}}\end{tabular}}}}%
    \put(0.48851276,0.12735609){\color[rgb]{0,0,0}\makebox(0,0)[lt]{\lineheight{17.25}\smash{\begin{tabular}[t]{l}{{\fontsize{11}{12}\selectfont {\textbf{\textcolor{white}{?}}}}}\end{tabular}}}}%
    \put(0.24004817,0.240700062){\color[rgb]{0,0,0}\makebox(0,0)[lt]{\lineheight{1.25}\smash{\begin{tabular}[t]{l}change lane and accelerate\end{tabular}}}}%
    %\put(0.55200183,0.03144499){\color[rgb]{0,0,0}\makebox(0,0)[lt]{\lineheight{1.25}\smash{\begin{tabular}[t]{l}\textit{sensor noise}\end{tabular}}}}%
    %\put(0.84772155,0.05641033){\color[rgb]{0,0,0}\makebox(0,0)[lt]{\lineheight{1.25}\smash{\begin{tabular}[t]{l}\textit{prediction}\end{tabular}}}}%
    %\put(0.84772155,0.02001033){\color[rgb]{0,0,0}\makebox(0,0)[lt]{\lineheight{1.25}\smash{\begin{tabular}[t]{l}\textit{space}\end{tabular}}}}%
  \end{picture}%
\endgroup%
}
  
  \vspace{-0.13cm}
  
  \resizebox{0.91\linewidth}{!}{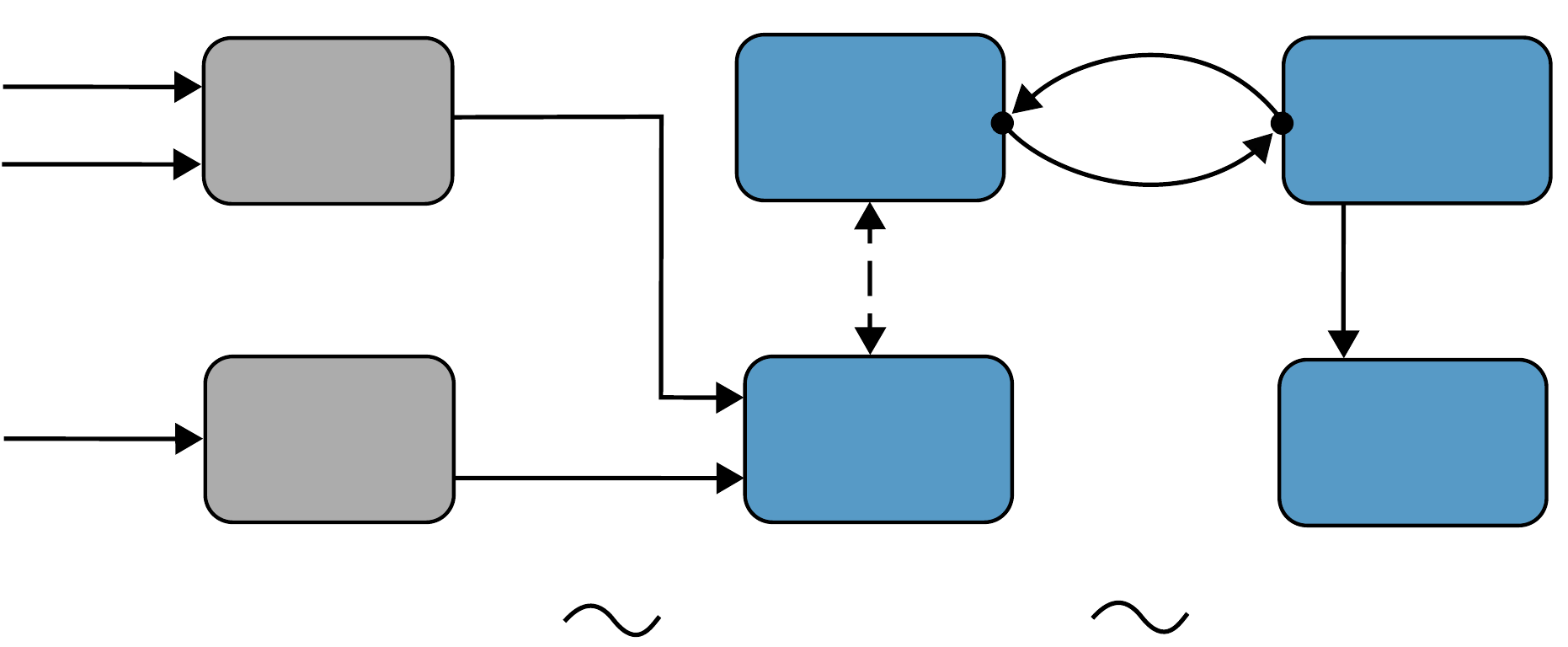}
  
  \vspace*{0.07cm}
  
  \caption{Risk Maps (RM) is implemented in a test vehicle and supports the driver in its behavior finding for a situation with an upcoming forced lane change. We focus on the modules indicated in blue. RM allows to visualize future risks and uncertainties for the driver.}

  \label{fig:demo_scenario}
\end{figure}

In previous research, we proposed the survival analysis \cite{puphal2019} for enhanced collision risk predictions under uncertainties. It is suitable for motion planning in situations with multiple interacting cars and has considerable advantages over common Time-To-X measures, which usually utilize time indicators as a risk approximation. In simulations \cite{puphal2018}, we proved that our method works reliably and does not produce collisions in a wide range of situations. 
For this paper, we target to extend and apply the risk framework on a test vehicle for online and predictive warning in forced lane changes (see Fig. \ref{fig:demo_scenario}). We thus show the benefitis of the approach: the easy visualization of the internal calculations.

Lane changes which are forced are complex and occur, e.g., due to road lanes that are ending, construction areas and parked cars. By evaluating behavior predictions with so-called Risk Maps (RM), we target to efficiently obtain lane change timings. A motion trajectory with minimal costs is selected (i.e., which balances risks with utility and comfort). In this process, a crucial component is the Relational Local Dynamic Map (R-LDM) that extracts driving paths, fuses the sensor information and prepares the driving situation. 

Essentially, in this paper, we move from simulation into the real world, which requires a prototype. 
The prototype consists of an inexpensive GNSS and multiple camera sensors. Fig. \ref{fig:demo_scenario}, at the bottom, shows the architecture of the system. 
In real experiments, RM with R-LDM will run on the test car and create warning outputs for the driver. The system can recommend the target velocities as well as path choices (go left, straight, etc.), in gap and no-gap variations. 
For the experiments, recordings are hereby drawn from a demonstration of the ADAS in the ITS European Congress 2019, described in \cite{vidas2021}.

To summarize, the contribution of this paper is threefold. On the scientific side, we build upon our previously published works \cite{puphal2019,puphal2018} and \cite{ldm2017}. However, we (i) extend the R-LDM to incorporate dynamic cars and their predicted behaviors. We show novel possibilities of how measurements and calculated variables can be saved in a database. Furthermore, we (ii) extend RM and plan besides longitudinal velocities, also lateral path choices, such as lane changes. A major novelty is the visualization of uncertainties in a predictive risk graph. On the technical side, we also propose a first, novel and real proof of concept. Concretely, we (iii) integrate both technologies and apply them on a prototype with an HMI system, highlighting the transparancy of the risk-based planner.

The remainder of this paper is structured as follows: Section \ref{subsec:rel} summarizes related research for ADAS. We continue with an R-LDM introduction in Section \ref{sec:fusion}, and outline the planning method of RM with Section \ref{sec:riskmaps}. Then, in Section \ref{sec:demo}, the test car and real-world demonstrations are outlined. Section \hspace{-0.05cm}\ref{sec:outl}\hspace{-0.05cm} finally presents conclusions\hspace{-0.01cm} plus\hspace{-0.022cm} future\hspace{-0.032cm} work. 

\subsection{Related Work}
\label{subsec:rel}

We will now compare the features of the risk-based ADAS with prior art from research. The focus of this section is put on the fields of trajectory prediction because the R-LDM uses map data to improve the predictions, on related risk methods, which are similar to the RM approach, and classical traffic models for motion planning in lane changes, since they are established baseline models. 

In the automotive field, road or map geometries are usually used to enhance the results of driving predictions for vehicles. For example, prevailing ADAS are extrapolating trajectories along paths from sensors and self-driving cars similarly rely on pre-recorded data \cite{ferguson2008}. As further alternatives to these map-based predictions, recently, learning algorithms are investigated. The authors of \cite{diehl2019}, e.g., learn typical trajectories in dynamic lane changes with Graph Neural Networks (GNN). Due to the vast behavior options for vehicles, prediction with fail safe backups are also shifting into focus \cite{pek2018}. 

In contrast to related work, we combine online sensor data with stored crowd-sourced map data. The data is managed by the R-LDM that allows for efficient data queries.

Alongside trajectory prediction, the development of driving risks\hspace{-0.03cm} becomes\hspace{-0.03cm} omnipresent\hspace{-0.03cm} in research. Hereby, risks are often sorted into discrete time or acceleration indicators, probabilis- tic risks as well as learned risks. Methods based on time metrics (see \cite{gassmann2019} for Time-To-Brake) convince due to their intuitiveness. However, neglecting uncertainty does not reproduce realistic driving situations. In probability considerations \cite{ruf2015}, tradeoffs must be found between accurate and computationally inexpensive models. Lastly, learning hazards \cite{watanabe2019} are powerful for specific complex events but struggle the most in untrained scenarios. 

The RM approach employs the survival analysis \cite{puphal2019}. On the one hand, this approach is more uncertainty-aware than state-of-the-art risks. On the other hand, the computational costs are reduced due to the analytical probability equations which form its basis. We probe situations with risks in the longitudinal (i.e., ego velocity) and lateral direction (i.e., possible lane change) to retrieve one optimal motion. In this context, probing is the driving prediction of a single ego behavior with an evaluation of induced risks with other cars.  

For further motion planning methods, we would eventually like to highlight traffic models in related work. They can easily be used for diverse applications. For instance, the Intelligent Driver Model (IDM) \cite{treiber2000} allows to follow preceding cars over kinematic equations. Using the extension of MOBIL \cite{kesting2007}, lane changes are considered in this process. The lane changes are based on simple predictions and are therefore understandable. However, the IDM does not regard driving uncertainties. 

In comparison, there are many advanced planners, such as lane change detectors \cite{zhang2020} or potential field approaches \cite{guo2019}, which are trying to include uncertainties in the car's positions for the planning. However, they often do not model uncertainty over the predicted time. Using RM, uncertainty-based risk is visualized over the predicted time and not only over the static positions. The proposed ADAS of this paper thus leverages the predictive nature in planning, while visualizing interpretable safety outputs to the user.

\vspace{-0.1cm}
\section{Representation of Driving Situations} 
\label{sec:fusion}
In support systems, clear structures of sensor measurements, trajectory patterns and driving situations are a prerequisite for their efficient real-world application. 
In this regard, world models are helpful as central storages, whereby so-called Local Dynamic Maps (LDM) are a potential way for their realization. LDM serve for the fusion of ADAS-relevant data.

According to the spirit of a Local Dynamic Map, we construct a database that represents several layers depending on their dynamicity. 
Dynamic entities, such as traffic participants, are stored on the highest layer.
The next layer is consisting of transient data, such as information on traffic signal phases, traffic congestion or road condition. 
On the second bottom layer, then, quasi-static objects, such as building outlines, are stored. The difference between the transient and quasi-static data is the dynamicity, which is changing on hours scale in the first and on the scale of days in the latter case.
Finally, the bottom layer serves to manage static data of the environment, which includes, e.g., spatial geometry data of lower layers that can be enriched by dynamic entities, e.g., traffic participants, of higher layers.

\begin{figure*}
    \centering
    \includegraphics[width=1.0\linewidth]{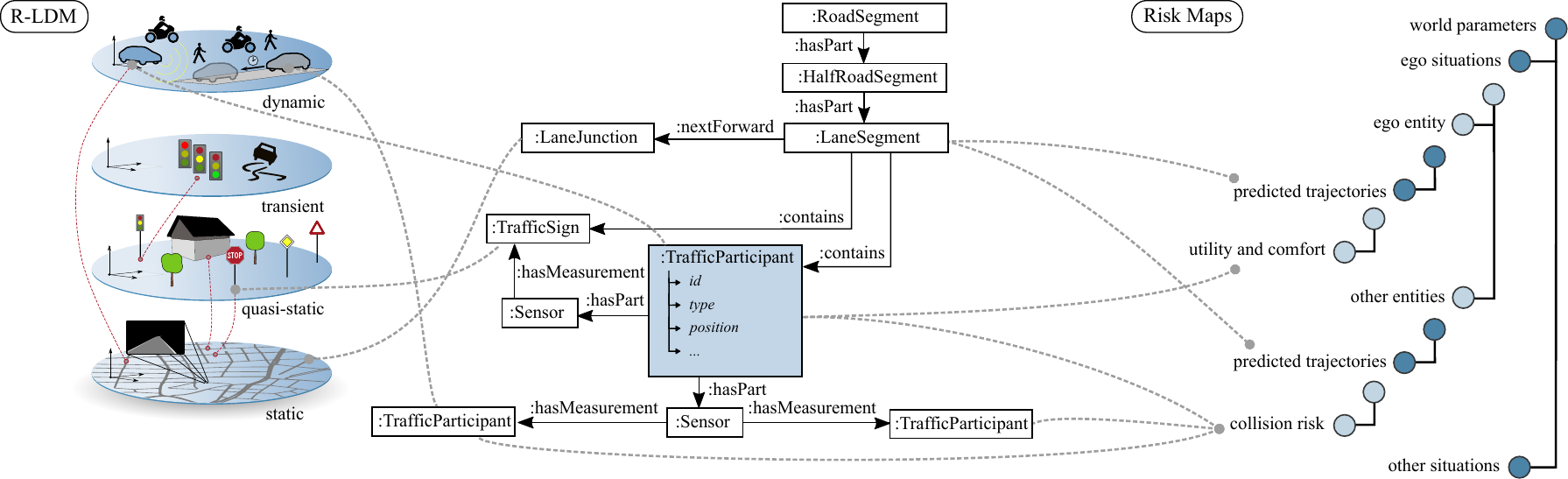}
    
    \vspace{0.13cm}
    
    \caption{Interaction of Relational Local Dynamic Map (R-LDM, left) and Risk Maps (right). Structured information, e.g., driving paths, from the local environment and traffic situation is leveraged for predicting trajectories, determining collision risk, utility and comfort. The data structures are connected in different levels of granularity, see dashed lines. Note that the parts of the figure can stand alone.}
    \label{fig:ldm_layers}
\end{figure*}

The left part of Fig. \ref{fig:ldm_layers} illustrates all of the four layers from the R-LDM concept. 
However, this paper focuses on the static and dynamic layer.
Our LDM implementation, the so-called Relational Local Dynamic Map (R-LDM), is hereby realized as a graph.
The R-LDM embodies an interconnected graph of nodes. 
Since relations present first order citizens of a graph, connections  between entities (e.g., spatial or temporal) can be described in a straightforward way. 
Attributes of nodes present an additional possibility to store information that is directly linked to a certain node.

This section builds upon this R-LDM, initially introduced in \cite{ldm2017}, and describes extensions towards the consideration of driving risks. 
We start with the description of static map data in Section \ref{subsec:map}, show how to manage dynamic entities with the graph in Section \ref{subsec:layers} and finish with the processing of sensor inputs in Section \ref{subsec:sensorproc}. 

Especially Section \ref{subsec:layers} and Section \ref{subsec:sensorproc} show how to connect trajectories as well as driving risks in the R-LDM, which was not described in the original R-LDM implementation, published in \cite{ldm2017}.

\subsection{Static Environment}
\label{subsec:map}
Tasks such as the estimation of collision risk strongly benefit from knowledge on the current and future, i.e., predicted, traffic situation. 
Such situation evolution can be described by a list of potential paths a driver can take.
Each path represents a sequence of positions that is derived from map data and builds the basis of the system's predictions.

The R-LDM enables the storage of road geometries on three levels of detail, i.e., road, half-road\footnote{Here, a half-road is defined to be the union of all lanes that point into the same direction.} and lane level. In terms of the graph, each level of detail is represented by a hierarchy of sub-graph patterns that consist of one central node (e.g., label ``:LaneSegment'') as well as related child nodes.
For example, a road is connected to up to two half-road nodes, while each half-road node can be connected to an arbitrary amount of lane nodes.

Going into further detail, the focus now lies on the lane-level geometry since it builds the basis of the path retrieval. Yet, the information applies to road and half-road level accordingly.
A :LaneSegment node stores the inherent information, such as a center polyline defining the direction of travel, as an attribute. 
Further node attributes include, amongst others,

\begin{itemize}
\item road type,
\item surface material,
\item lane marking type and
\item road curvature.
\end{itemize}

Subordinate entities, e.g., lane markings or boundaries detected by sensors, are represented by individual nodes and are connected via ``:hasPart'' relations. 
Further connected nodes include traffic signs as well as further entities of the same or different R-LDM layers that are related in a direct or indirect sense. 

The list of attributes and the level of detail 
can be chosen in accordance with the individual use case. 
However, also the omission of information can be beneficial, since the computing time correlates with the amount of data stored in the graph.  

Regarding the acquisition of raw map geometry data, common approaches are the extension of publicly available sources (e.g., OpenStreetMaps \cite{OSM}), the parsing of digital orthophotos (satellite imagery) or the derivation of road geometry from pre-recorded GNSS trajectories. 
Assuming that a vehicle drives in the center of the road, the latter approach natively presents a suitable means for validating the hypothesis.

\subsection{Dynamic Environment}
\label{subsec:layers}

As already mentioned, a crucial group of objects are the traffic participants, such as, vehicles, cyclists, and pedestrians. This is especially the case for a predictive warning system for dynamic lane changes.
In this context, the R-LDM can be interpreted as an ego-centered knowledge hub that represents the surrounding of an agent zero.
More specifically, the agent zero is moving within a (potentially) pre-mapped environment that is continuously enriched by measurements and sensed data, which is shown in the middle of Fig. \ref{fig:ldm_layers}. 

The ego car, or more general the ego vehicle, is represented by a node.
For any type of sensor, a node with the label :Sensor is created that is connected to the entity node via a ``:hasPart'' link. 
The type of sensor is stored as an attribute with the key type.
Depending on type and amount of information, the actual data can be stored explicitly as attribute or implicitly  as a reference to another place as, e.g., a hard drive or a time series database.  
In the case of explicit storage, it can be inefficient and computationally expensive to push data from an Inertial Measurement Unit (IMU) with 100 Hz. 
Instead, the sensor data updates are pushed at a pre-defined, suitable frequency to the R-LDM.

Any kind of measurement, e.g., object detection from cameras or GNSS position estimates can be considered, including other traffic participants.  
Each sensed entity, such as another vehicle, can afterwards be represented by a graph node that is connected via a ``hasMeasurement'' link. 

Additionally, traffic participants are also connected to static elements, in other words, the road infrastructure.
One common example is a lane segment that is utilized by vehicles, indicated in the R-LDM by the ``:contains'' links between the lane and vehicle nodes. 
In this context, the center of a lane serves as an approximation for the driving path, i.e., the path that is most likely be driven by the vehicle.
In multi-lane scenarios, several paths can be determined with discrete sets of behavior, e.g., stay on the lane or perform a lane change. 

While a path represents consecutive points in a 2D-space, a trajectory also conveys timestamps for each data point.  

For Risk Maps (RM), we require trajectories that  are based on the assumption of fixed dynamics (e.g., constant velocity) for predicted times and are described as a sequence of position $(\text{x}, \text{y})$, velocity $v$, acceleration $a$ and jerk $j$. 
Furthermore, RM considers benefits of the ego vehicle (e.g., desired speed) and risks between the ego and other vehicles. 
Fig. \ref{fig:ldm_layers}, on the right, depicts how RM benefits from the R-LDM.

\subsection{Data Processing} 
\label{subsec:sensorproc}

\begin{figure}[t!]
  \centering
  
  \vspace{0.18cm}
  
  \resizebox{0.95\linewidth}{!}{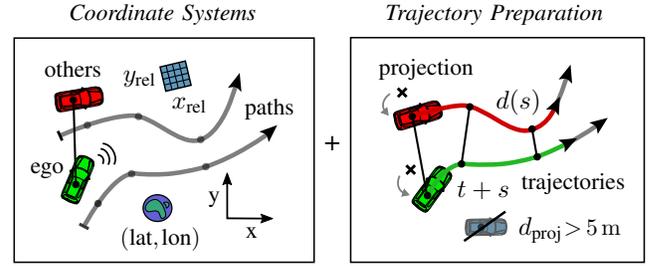}
  \vspace*{0.22cm}
  \caption{Pre-processing steps for map-based planning in R-LDM. Left: Car signals are transformed into the same global reference frame. Right: For motion predictions, we filter other cars that are far away.} 
  \label{fig:coord_trans}
\end{figure}

The preceding Sections \ref{subsec:map} and \ref{subsec:layers} explained how map data and dynamic entities can be stored in a graph. 
Now, we illustrate how stored data interacts with acquired sensor data. In this context, Fig. \ref{fig:coord_trans} shows the ego (green) and a sensed (red) vehicle that are both projected onto the closest map paths. 
For doing so, two techniques are employed to process such sensor signals: 1) the alignment of the ego car on a driving path and 2) the filtering of unlikely obstacle detections. 
In this way, we can afterwards predict the trajectories along map paths. 

\subsubsection{Ego Alignment}
The positions $(\text{lat}, \text{lon})$ of the ego car from a GNSS sensors are given.
In a first step, we transform the GNSS signals to Cartesian coordinates. 
The equirectangular projection with a radius of earth $r_e$ and latitude $\text{lat}_0$, which is close to the center of the map, allows for the conversion of geodetic GNSS coordinates $(\text{lat}, \text{lon})$ to Cartesian coordinates $(\text{x}, \text{y})$ via the equations
\begin{align}
\text{x} = r_e &\cos \text{lat}_0 \cdot \text{lon}, \\ %\vspace{0.2cm}
\text{y} &= r_e \cdot \text{lat}.
\end{align}

Here, the $x$-axis of the global reference frame is facing east, while $y$ is facing north. 
We project the ego car on the closest  path. 
This description assumes the earth to be an evenly round 

\noindent globe. However, the reader can choose any earth model that fits the individual accuracy requirements, such as assuming an earth ellipsoid.  
For deeper insight into map-based lane-level localization, we refer to \cite{localization2020}.

\subsubsection{Obstacle Filter}
The visual detection of the surrounding vehicles requires a coordinate transformation of their positions from an ego-relative body frame $(x_{\text{rel}}, y_{\text{rel}})$ into the shared world reference frame $(x, y)$. 
For this purpose, we use a rotation transformation with the angle $\Theta$ between both reference frame. Hereby, we use the standard, right-handed orientation of the world frame. If we additionally assume flat surfaces, we get

\vspace{-0.44cm}

\begin{equation}
\text{x} = \cos \Theta \cdot x_{\text{rel}} - \sin \Theta \cdot y_{\text{rel}},
\end{equation}

\vspace*{-0.47cm}

\begin{equation}
\text{y} = \sin \Theta \cdot x_{\text{rel}} + \cos \Theta \cdot y_{\text{rel}}.
\end{equation}

\vspace{0.02cm}

\noindent The equations can be extended to cover 3D-space in a straightforward way. This allows for the consideration of, e.g., non-flat environments. 

Similar to the ego alignment on its path, the other cars are projected to their paths as well. However, the main difference is that we only consider vehicles that are close to any path and filter cars with a distance $d_{\text{proj}}>\unit[5]{m}$ away. This geofencing limits position errors from sensors. In the end, we compute the distance $d$ in the global frame for the collision risk calculation. 

\noindent In summary, the R-LDM allows us to prepare situations for risk probing within RM. This method will be described in the following. 

\section{Probing inside Risk Maps}
\label{sec:riskmaps}
\vspace{0.02cm}
Using the R-LDM, we are able to fuse measured data into driving situations. 
However, assessing a multi-lane situation based on its risks thoroughly and, at the same time, in a fast manner remains challenging. The reason lies mostly in the variety of possible predictions for such situations. Additionally, there are underlying driving uncertainties, which arise from car sensors and likewise, e.g., from the driver behavior. 

In this context, planning methods must determine an optimal maneuver for the ego driver. With map data, the driving space was constrained and we differentiated between static paths on the one hand, and dynamic trajectories on the other hand. Still, a single path and trajectory that is safe and beneficial for the ego driver needs to be found. 
Therefore, so-called "probing" inside RiskMaps (RM) will be leveraged. We consider probing as the prediction of an ego driving motion and its evaluation of induced risks. By intelligently probing adequate numbers of fixed trajectories along paths in RM, motions will be efficiently planned. 

This section describes the concepts behind the planner RM. For this purpose, we will first explain the trajectory probing in Section \ref{subsec:traj} with RM and we will continue with the cost evaluation of risk and benefits in Section \ref{subsec:survival}. In the last Section \ref{subsec:lat} and Section \ref{subsec:warning}, we will finalize the description of RM with the extension of path probing, which ultimately allows to warn the driver. 

\begin{figure}[t!]
  \centering
  \vspace{0.06cm}
  \resizebox{0.97\linewidth}{!}{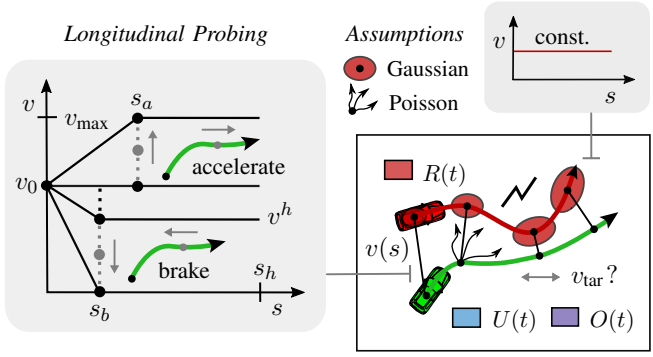}
  \vspace{0.25cm}
  \caption{Trajectory probing within RM. Left: Ego trajectories are planned with different acceleration and braking strengths. Right: Evaluation of the risks, utilities and comforts for constant velocity of another car. The figure shows the risk calculation in more detail.}
  \label{fig:traj_variation}
\end{figure} 

\vspace{0.05cm}
\subsection{Trajectory Probing}
\label{subsec:traj}
Hereinafter, we analyze a car-to-car encounter with collision risks, depicted in Fig. \ref{fig:traj_variation}. For every ego trajectory variation $h$, constant velocity is assumed for the other car (i.e., $v=\text{const.}$). For the ego car, we predict and create variations for a piece-wise function composed of an acceleration/braking phase and a zero-acceleration, resp., constant speed segment. Technically, we first sample $N_t$ velocity profiles $v(s)$ on the path that start at the current velocity $v_0$ and end at the planned velocities $v^h$ in the future time $s$, equidistantly sampled inside $v\in[0, v_{\text{max}}]$. In total, we thus get

\vspace{-0.02cm}

\begin{equation}
v^{h} = \frac{h}{N_t-1} \cdot v_{\text{max}} \text{\hspace{0.02cm} with \hspace{0.025cm}} h \in 0, ..., N_t-1.
\end{equation}

\vspace{0.05cm}

To reach those end velocities $v^h$, the accelerations $a^h$ are afterwards calculated that are maximal with $a_{\text{max}}$ in case of an ego trajectory that corresponds to $v_{\text{max}}$. For the cases ending in a full stop $v^h=\unit[0]{m/\text{sec}}$ of the ego vehicle, the acceleration $a^h$ becomes minimal with $a_{\text{min}}$. Note that $a_{\text{min}}$ represents hereby a negative acceleration (i.e., braking motion). This leads to

\vspace{-0.5cm}

\begin{align}
\text{if } v^h > v_0\text{: \hspace{0.025cm}} a^{h} = a_{\text{max}} \cdot \frac{v^h-v_0}{v_{\text{max}}- v_0}, \\
\text{else if } v^h < v_0\text{: \hspace{0.025cm}}a^{h} = a_{\text{min}} \cdot \frac{v_0-v^h}{v_0}. 
\end{align}

\vspace{0.01cm}

The intervals with either an acceleration or braking phase are planned for the predicted durations of $s\in[0, s_a]$ or $s\in[0, s_b]$, respectively. We formulate the included times as

\begin{equation}
s_{a} = \frac{v_0}{a_{\text{max}}} \text{\hspace{0.015cm} and \hspace{0.01cm}} s_{b} = |\frac{v_0}{a_{\text{min}}}|. %\left\| \right\|
\end{equation}

\vspace{0.1cm}

\noindent The included time in the equations is denoted as $s$ since it is not the real time $t$ but the predicted time. We assume a constant velocity for the ego vehicle in the subsequent segment $[s_a, s_h]$ and $[s_b, s_h]$. The parameter $s_h$ defines the time horizon of the velocity profile and is set according to the task. Fig. \ref{fig:traj_variation} (on the left) depicts this longitudinal probing. 

In the presented planner, a target velocity $v^h = v_{\text{tar}}$ from the RM set is eventually selected based on the explicit tradeoffs between risks $R(t)$ and benefits, which are further divided into the ego utility $U(t)$ and ego comfort $O(t)$, see Fig. \ref{fig:traj_variation} on the right. In this process, at least $N_t\geq3$ ego trajectories have to be sampled so that RM can choose between an acceleration and a braking option.  
Altogether, RM could potentially represent a fast planner applicable to avoid accidents. 

The next section will now describe the mentioned costs to select an optimal motion. Using the survival analysis approach, we include probabilistic assumptions of Gaussian uncertainty (visualized as 2D, red ellipses) and of Poisson uncertainty (escape arrows for the ego cars). However, this section represents a summary of the approach. For more details, please refer to the original publication \cite{puphal2019}. We extend and show novel ways how to plan lane changes afterwards.

\vspace{-0.2cm}
\subsection{Survival Analysis}
\label{subsec:survival}

We cannot presume that predictions of other cars, e.g., constant velocity, will be followed in reality. The survival analysis thus models uncertainties for their predicted positions with 2D Gaussians that are growing over time. These uncertainties are influenced by sensor uncertainty (dominant in first steps) and behavior uncertainty (dominant in later steps). Finally, a collision probability is given by the overlap of the Gaussians from the ego vehicle and all other vehicles.\footnote{Note that this collision probability is, in the end, dependent on the relative velocity and distance of the vehicles.} 

Since risk is defined as the probability of this critical event multiplied with a damage outcome, we include a severity term in the final risk output. 
Besides collision risks, RM allow for the analysis of further critical events, such as upcoming sharp curves. More specifically, RM model the probability to lose control and skid off during lane changes. In these cases, we assume 1D Gaussians around the ego car and look at the lateral acceleration which is influenced by road curvatures.

\vspace{0.015cm}
\subsubsection{Risk Equations}
With the help of a Poisson process, we are able to accumulate probabilities over the future time $s$ (i.e., from collisions and a curve) with the damages into a single, scalar risk value $R(t)$ for the current time  $t$. In detail, we calculate the time difference between two critical events $\tau_{\text{crit}}^{-1}(s)$. Those critical events are then divided into collision rates $\tau_{\text{coll},j}^{-1}$ (with the index defining the considered other car $j$) and a curve rate $\tau_{\text{curv}}^{-1}$. 

Due to the included survival function $S$ in a Poisson process, risks that occur further away in the prediction are considered less in the risk $R(t)$. This so-called ``escape'' effect is modeled for the ego car. We thus describe $R(t)$ by 

\vspace{-0.2cm}

\begin{equation}
R(t) = \int_0^{\infty} (\sum_j \tau_{\text{coll},j}^{-1}D_{\text{coll},j}+\tau_{\text{curv}}^{-1}D_{\text{curv}})S \,ds.
\label{eq:risk}
\end{equation}  
For Eq. (\ref{eq:risk}), the severity is hereby formulated in a desired accuracy based on a collision model $D_{\text{coll}}$ and $D_{\text{curv}}$. 

\begin{figure}[t!]
  \centering
  
  \vspace*{-0.252cm}
  
  \resizebox{0.82\linewidth}{!}{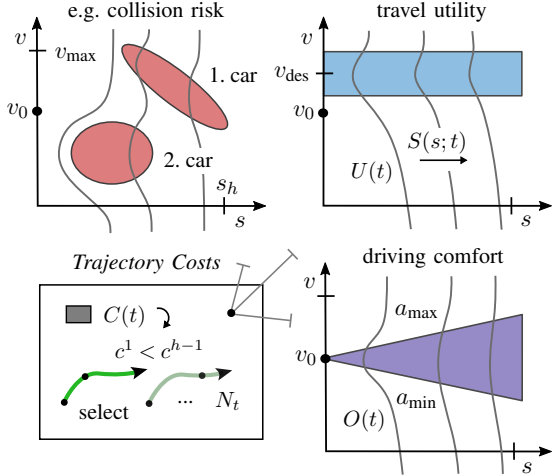}
  \vspace{0.1cm}
  \caption{Visualization of the driving risk (i.e., from collision and curve), utility and comfort costs. The trajectory costs are computed over the future time and then integrated with the survival analysis to obtain a single cost scalar. Note: The graphs should only give a qualitative notion of how the costs may look like.} 
  \label{fig:single_costs}
\end{figure} 

In the end, we can visualize a risk graph showing risk hot spots for the taken velocity profile that correspond to the other cars. Such a risk graph is plotted in Fig. \ref{fig:single_costs} on the top left. As an example, it shows risk spots for two separate cars, which become visible for trajectories when the ego car come closest to the other car, while the shape of the hot spots depends on the Gaussian uncertainties of the vehicles. At this point, it should be highlighted that the risk graph shows risks over the future time, which is different to common potential field visualizations, such as \cite{guo2019}.

\vspace{0.03cm}
\subsubsection{Utility and Comfort Prediction}

The target of RM is to mimize risks as well as to maximize benefits $B(t)$. The total

\noindent costs are calculated with $C(t) = R(t)-B(t) = R(t)-U(t)-O(t)$.
In this sense, we express benefits with high utility and comfort, both also depicted in Fig. \ref{fig:single_costs}. 

Utility $U(t)$ depends on the driven ego velocity $v_1$, which attempts to reduce the general required time to arrive at a goal. Additionally, we consider a desired velocity $v_d$ for individual preferences. 
For decreasing effects of $U(t)$ at higher times $s$, we multiply the components with the survival function $S$ and write 

\vspace{-0.25cm}

\begin{equation}
U(t) = \int_0^{\infty} (b^t |v_1| + b^d |v_1 - v_d|) S \,ds.
\label{eq: utility}
\end{equation} 

\vspace{0.02cm}

Comfort $O(t)$ takes ego acceleration $a_1$ and jerk $j_1$ into account and ensures less abrupt switches between selected behaviors. The major target is still to avoid risks. Consequently, comfort is also reduced over the survival function $S$. For $O(t)$, we gain
\vspace{-0.1cm}
\begin{equation}
O(t) = \int_0^{\infty} -(b^c |a_1| + b^j |j_1|) S \,ds.
\label{eq: comfort}
\end{equation} 

With the weight parameters $b^t$, $b^d$, $b^c$ and $b^j$, in both $U(t)$ and $O(t)$, the importance of risk versus benefit can be tuned. This enables us to select a single trajectory with minimal costs $c^h$ among all sampled trajectories $N_t$. Particularly, in Fig. \ref{fig:single_costs}, the first trajectory $v^1$, i.e., with $h=1$, was chosen because $c^1$ represents the smallest cost of all trajectories. With the cost graphs of $U(t)$ and $O(t)$, visualized also in the figure alongside risks $R(t)$, we can therefore intuitively indicate the underlying reason for any selection of the planner.

\subsection{Path Probing} 
\label{subsec:lat}

For the last module of the resulting warning system, this section outlines path planning. Building upon the ego trajectory probing and cost evaluation from RM, a novel path probing technique is proposed for RM. In this way, the forced lane change of the introduction is solvable. The driver has the support of the HMI, recommending a target motion. 

\begin{figure}[t!]
  \centering
  \vspace{0.23cm}
  \resizebox{0.89\linewidth}{!}{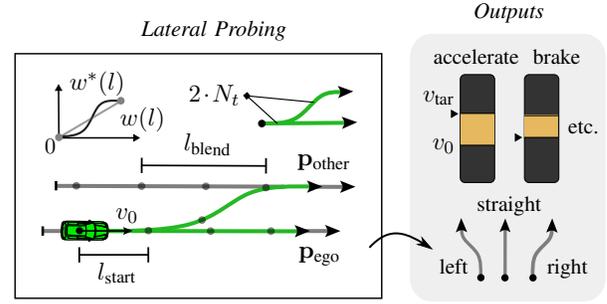}
  \vspace{0.15cm}
  \caption{Signal outputs of RM. By also probing the costs for possible ego paths, we can plan tactical lane changes. Left: Blending of path options and their selection. Right: HMI with target velocity and lane change advice.}
  
  \vspace{0.05cm}
  
  \label{fig:path_blending}
\end{figure}

As already mentioned, in large multi-lane roads (i.e., highways), a driver has distinct spatial options with the possibility of performing a lane change. A forced lane change arises either for, e.g., following the navigation route, avoiding neighboring cars or both at the same time. In these instances, the lane change path and its start time and duration must be determined. The RM make tactical decisions, which are, e.g., lane changes for a future time (depicted in Fig. \ref{fig:path_blending}). 

In detail, we compute a path change that serves to blend between centerlines of the ego lane and of a neighbouring lane. The blending begins at a longitudinal distance $l_{\text{start}}$, which is defined as
\vspace{-0.06cm}
\begin{equation}
l_{\text{start}} = v_0 \cdot s_{\text{start}}.
\end{equation}
Accordingly, this distance $l_{\text{start}}$ depends on the current velocity $v_0$ and the future time $s_{\text{start}}$. The length of the blending interval $l_{\text{blend}}$ is, in contrast, influenced by the current, lateral distance $d_{\text{path}}$ between the two path options. Since the segment end $l_{\text{end}}$ is composed of $l_{\text{start}}$ and $l_{\text{blend}}$, we retrieve
\vspace{0.03cm}
\begin{equation}
l_{\text{end}} =  l_{\text{start}} + l_{\text{blend}} = l_{\text{start}} + v_0\sqrt{l_c \cdot d_{\text{path}}}.  
\label{eq:path_length}
\end{equation}

\vspace{0.03cm}

\noindent In Eq. (\ref{eq:path_length}), the parameter $l_c$ defines the scale factor for the increase. Simply put, for higher $l_c$, the lane change would also take more time.

In a final, next step, we blend the ego path $\mathbf{p}_{\text{ego}}(l)$ into the neighboring path $\mathbf{p}_{\text{other}}(l)$ within the segment $l_{\text{blend}}$, using a sigmoidal weighting term $w^*(l)$. Paths from Section~\ref{subsec:map} are initially resampled with evenly spaced points. The computed lane change path $\mathbf{p}_{\text{blend}}(l)$ then follows, with 
\vspace{0.07cm}
\begin{equation}
\mathbf{p}_{\text{blend}}(l) = (1-w^*(l)) \cdot \mathbf{p}_{\text{ego}}(l) + w^*(l) \cdot \mathbf{p}_{\text{other}}(l). 
\end{equation}

\vspace{0.08cm}

\noindent Generally, linear blending is achieved via the term
\vspace{0.05cm}
\begin{equation}
w(l) = \frac{l - l_{\text{start}}}{l_{\text{blend}}}, \text{with } l\in[l_{\text{start}}, l_{\text{end}}].
\end{equation}

\vspace{0.05cm}

\noindent and we thus utilize the weight $w(l)$ in a sigmoidal weight function $w^*(l)$ to gain
\vspace{0.05cm}
\begin{equation}
\vspace*{-0.1cm}
w^*(l) = \frac{{\fontsize{6}{8}\selectfont 1}}{1+e^{\mbox{\fontsize{9}{9.5}\selectfont{$-k(w(l)+0.5$)}}}}.
\vspace*{0.19cm}
\end{equation}
The constant $k$ allows to hereby tune the blending steepness and should be set to achieve a smooth blending. 
\vspace{0.05cm}

Fig. \ref{fig:path_blending} (left box) depicts two prototypical paths. They are calculated with the aforesaid variables. Note that points in the path have to be sampled densely enough for a reasonable blending. With the path blending, we can initiate an immediate lane change by setting $s_{\text{start}}=\unit[0]{\text{sec}}$, while a tactical change is obtained with, e.g., $s_{\text{start}}\hspace{-0.02cm}=\hspace{-0.02cm}\unit[2]{\text{sec}}$. A utility offset is necessary for the costs in order to incentivize the planner making a lane change. After all, RM choose the path based on its costs. 

\subsection{Selection and Warning}
\label{subsec:warning}

For multiple paths, the path blending is done iteratively. With the total path number $M_p$, ultimately, $M_{\text{p}} \cdot N_{\text{t}}$ samples are generated in RM, since ego trajectories are varied on each path. We select a single trajectory with the lowest costs for obtaining the target velocity $v_{\text{tar}}$ as well as the target path $\mathbf{p}_{\text{tar}}$. In other words, RM probe in longitudinal and lateral directions. The runtime of RM is constant because we have a fixed sample size. This suits very well for real-time purposes. In the case of Fig. \ref{fig:path_blending}, we get $M_p=2$ and sample $2 \cdot N_{\text{t}}$ trajectories. 

The planned, safe motion is now transferred into a driver suggestion or warning in an HMI so that it is explained to the driver. 
We compare the planned with the actual one to infer a warning. 

As shown in Fig. \ref{fig:path_blending} (right-hand side), the developed HMI contains a velocity scale with the current velocity $v_0$ and the safe velocity $v_{\text{tar}}$. Depending on the difference $|v_0-v_{\text{tar}}|$, the driver needs to change its behavior (accelerate, brake, etc.). Furthermore, a directional arrow depicts the path choice from the planner, which can be either ``left'', ``straight'' or ``right''. The driver should make a lane change, i.e., left, right, or stay on the lane, i.e., straight motion. The figure shows some output examples of the HMI, which are derived from the solution of the planner.
\vspace{0.12cm}

\section{Real-World Demonstration}
\label{sec:demo}
We described the concepts behind R-LDM (Relational Local Dynamic Map) as a supporting module for a risk-based motion planning, and the Risk Maps (RM) technology representing an uncertainty-aware planner determining optimal driving trajectories and paths.

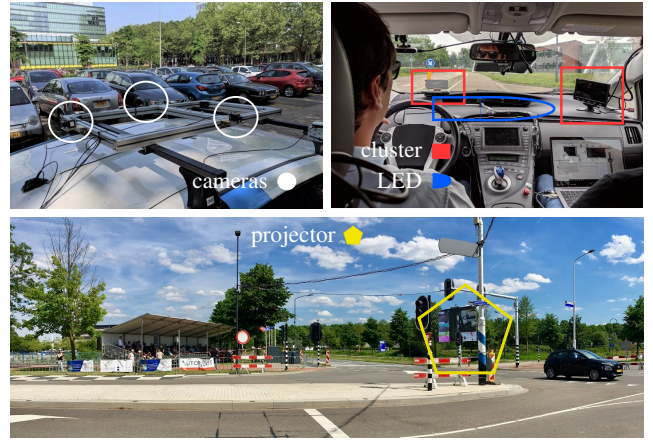
\begin{figure}[!t]
  \centering
    \vspace{-0.25cm}
    \resizebox{0.4674\linewidth}{!}{%% Creator: Inkscape inkscape 0.92.4, www.inkscape.org
%% PDF/EPS/PS + LaTeX output extension by Johan Engelen, 2010
%% Accompanies image file 'vicomtech_cams3.pdf' (pdf, eps, ps)
%%
%% To include the image in your LaTeX document, write
%%   \input{<filename>.pdf_tex}
%%  instead of
%%   \includegraphics{<filename>.pdf}
%% To scale the image, write%% Creator: Inkscape inkscape 0.92.4, www.inkscape.org
%% PDF/EPS/PS + LaTeX output extension by Johan Engelen, 2010
%% Accompanies image file 'vicomtech_cams.pdf' (pdf, eps, ps)
%%
%% To include the image in your LaTeX document, write
%%   \input{<filename>.pdf_tex}
%%  instead of
%%   \includegraphics{<filename>.pdf}
%% To scale the image, write
%%   \def\svgwidth{<desired width>}
%%   \input{<filename>.pdf_tex}
%%  instead of
%%   \includegraphics[width=<desired width>]{<filename>.pdf}
%%
%% Images with a different path to the parent latex file can
%% be accessed with the `import' package (which may need to be
%% installed) using
%%   \usepackage{import}
%% in the preamble, and then including the image with
%%   \import{<path to file>}{<filename>.pdf_tex}
%% Alternatively, one can specify
%%   \graphicspath{{<path to file>/}}
%% 
%% For more information, please see info/svg-inkscape on CTAN:
%%   http://tug.ctan.org/tex-archive/info/svg-inkscape
%%
\begingroup%
  \makeatletter%
  \providecommand\color[2][]{%
    \errmessage{(Inkscape) Color is used for the text in Inkscape, but the package 'color.sty' is not loaded}%
    \renewcommand\color[2][]{}%
  }%
  \providecommand\transparent[1]{%
    \errmessage{(Inkscape) Transparency is used (non-zero) for the text in Inkscape, but the package 'transparent.sty' is not loaded}%
    \renewcommand\transparent[1]{}%
  }%
  \providecommand\rotatebox[2]{#2}%
  \newcommand*\fsize{\dimexpr\f@size pt\relax}%
  \newcommand*\lineheight[1]{\fontsize{\fsize}{#1\fsize}\selectfont}%
  \ifx\svgwidth\undefined%
    \setlength{\unitlength}{135.16000366bp}%
    \ifx\svgscale\undefined%
      \relax%
    \else%
      \setlength{\unitlength}{\unitlength * \real{\svgscale}}%
    \fi%
  \else%
    \setlength{\unitlength}{\svgwidth}%
  \fi%
  \global\let\svgwidth\undefined%
  \global\let\svgscale\undefined%
  \makeatother%
  \begin{picture}(1,0.74932613)%
    \lineheight{1}%
    \setlength\tabcolsep{0pt}%
    \put(0,0){\includegraphics[width=\unitlength,page=1]{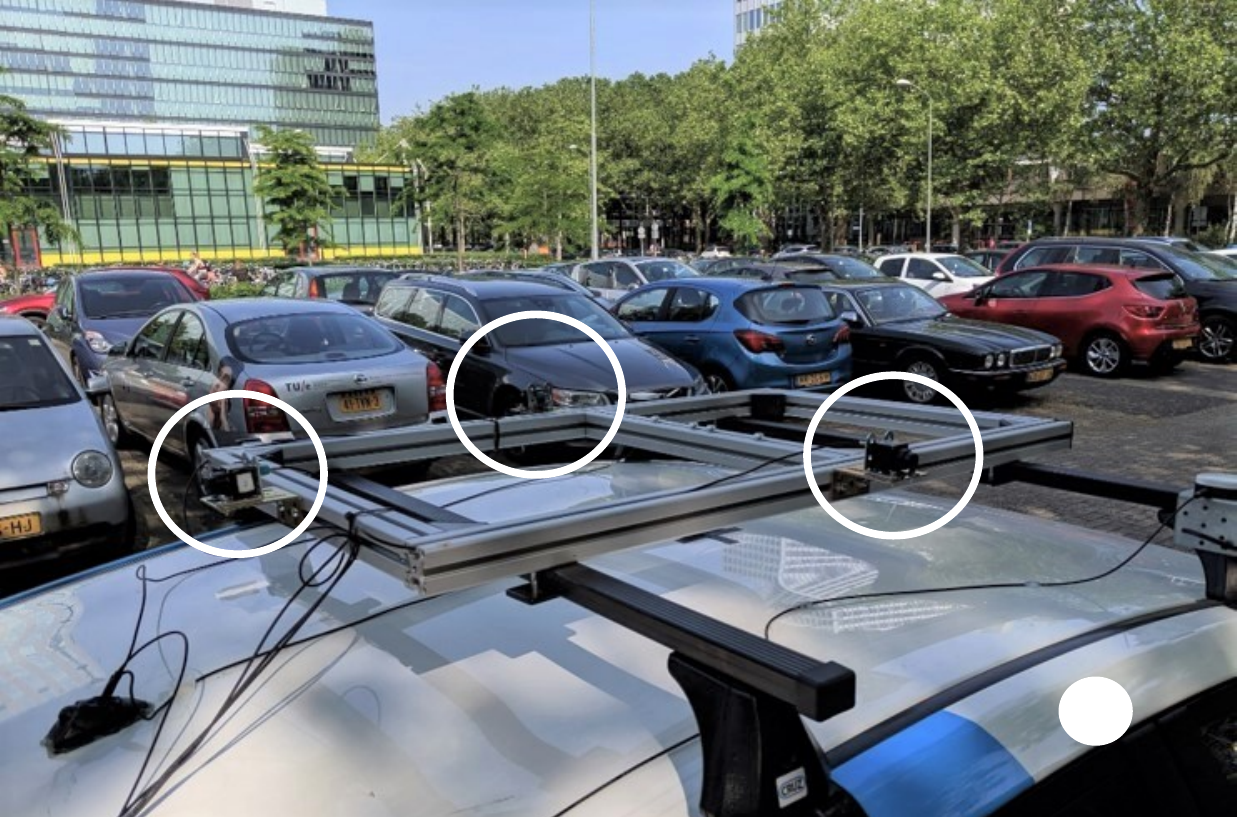}}%
    \put(0.58297584,0.06408937){\color[rgb]{0,0,0}\makebox(0,0)[lt]{\lineheight{1.25}\smash{\begin{tabular}[t]{l}\textcolor{white}{cameras}\end{tabular}}}}%
  \end{picture}%
\endgroup%
}
    \resizebox{0.4674\linewidth}{!}{%% Creator: Inkscape inkscape 0.92.4, www.inkscape.org
%% PDF/EPS/PS + LaTeX output extension by Johan Engelen, 2010
%% Accompanies image file 'vicomtech_hmi.pdf' (pdf, eps, ps)
%%
%% To include the image in your LaTeX document, write
%%   \input{<filename>.pdf_tex}
%%  instead of
%%   \includegraphics{<filename>.pdf}
%% To scale the image, write
%%   \def\svgwidth{<desired width>}
%%   \input{<filename>.pdf_tex}
%%  instead of
%%   \includegraphics[width=<desired width>]{<filename>.pdf}
%%
%% Images with a different path to the parent latex file can
%% be accessed with the `import' package (which may need to be
%% installed) using
%%   \usepackage{import}
%% in the preamble, and then including the image with
%%   \import{<path to file>}{<filename>.pdf_tex}
%% Alternatively, one can specify
%%   \graphicspath{{<path to file>/}}
%% 
%% For more information, please see info/svg-inkscape on CTAN:
%%   http://tug.ctan.org/tex-archive/info/svg-inkscape
%%
\begingroup%
  \makeatletter%
  \providecommand\color[2][]{%
    \errmessage{(Inkscape) Color is used for the text in Inkscape, but the package 'color.sty' is not loaded}%
    \renewcommand\color[2][]{}%
  }%
  \providecommand\transparent[1]{%
    \errmessage{(Inkscape) Transparency is used (non-zero) for the text in Inkscape, but the package 'transparent.sty' is not loaded}%
    \renewcommand\transparent[1]{}%
  }%
  \providecommand\rotatebox[2]{#2}%
  \newcommand*\fsize{\dimexpr\f@size pt\relax}%
  \newcommand*\lineheight[1]{\fontsize{\fsize}{#1\fsize}\selectfont}%
  \ifx\svgwidth\undefined%
    \setlength{\unitlength}{135.16000366bp}%
    \ifx\svgscale\undefined%
      \relax%
    \else%
      \setlength{\unitlength}{\unitlength * \real{\svgscale}}%
    \fi%
  \else%
    \setlength{\unitlength}{\svgwidth}%
  \fi%
  \global\let\svgwidth\undefined%
  \global\let\svgscale\undefined%
  \makeatother%
  \begin{picture}(1,0.66119441)%
    \lineheight{1}%
    \setlength\tabcolsep{0pt}%
    \put(0,0){\includegraphics[width=\unitlength,page=1]{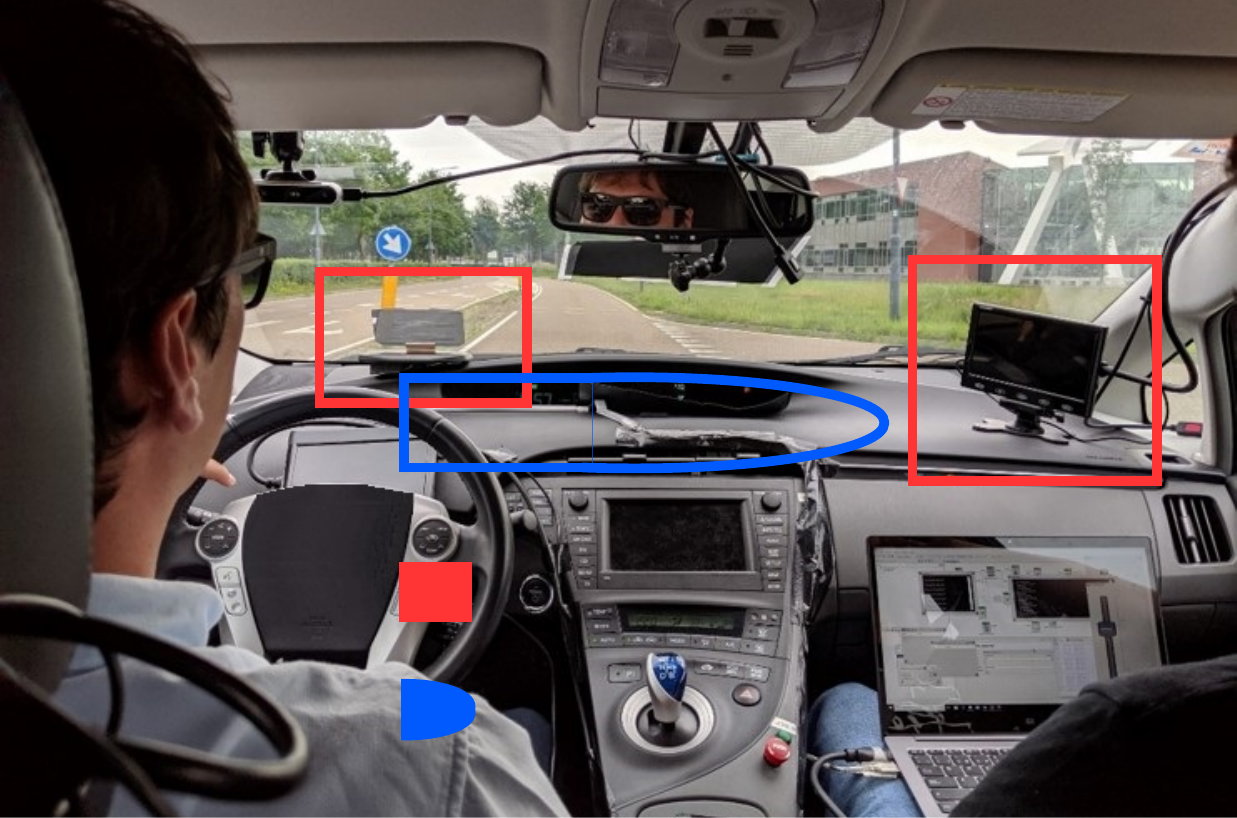}}%
    \put(0.09480787,0.16345403){\color[rgb]{0,0,0}\makebox(0,0)[lt]{\lineheight{1.25}\smash{\begin{tabular}[t]{l}\textcolor{white}{cluster}\end{tabular}}}}%
    \put(0.14680787,0.06307626){\color[rgb]{0,0,0}\makebox(0,0)[lt]{\lineheight{1.25}\smash{\begin{tabular}[t]{l}\textcolor{white}{LED}\end{tabular}}}}%
  \end{picture}%
\endgroup%
}
    
    \vspace*{0.13cm}
    \resizebox{0.95\linewidth}{!}{%% Creator: Inkscape inkscape 0.92.4, www.inkscape.org
%% PDF/EPS/PS + LaTeX output extension by Johan Engelen, 2010
%% Accompanies image file 'vidas.pdf' (pdf, eps, ps)
%%
%% To include the image in your LaTeX document, write
%%   \input{<filename>.pdf_tex}
%%  instead of
%%   \includegraphics{<filename>.pdf}
%% To scale the image, write
%%   \def\svgwidth{<desired width>}
%%   \input{<filename>.pdf_tex}
%%  instead of
%%   \includegraphics[width=<desired width>]{<filename>.pdf}
%%
%% Images with a different path to the parent latex file can
%% be accessed with the `import' package (which may need to be
%% installed) using
%%   \usepackage{import}
%% in the preamble, and then including the image with
%%   \import{<path to file>}{<filename>.pdf_tex}
%% Alternatively, one can specify
%%   \graphicspath{{<path to file>/}}
%% 
%% For more information, please see info/svg-inkscape on CTAN:
%%   http://tug.ctan.org/tex-archive/info/svg-inkscape
%%
\begingroup%
  \makeatletter%
  \providecommand\color[2][]{%
    \errmessage{(Inkscape) Color is used for the text in Inkscape, but the package 'color.sty' is not loaded}%
    \renewcommand\color[2][]{}%
  }%
  \providecommand\transparent[1]{%
    \errmessage{(Inkscape) Transparency is used (non-zero) for the text in Inkscape, but the package 'transparent.sty' is not loaded}%
    \renewcommand\transparent[1]{}%
  }%
  \providecommand\rotatebox[2]{#2}%
  \newcommand*\fsize{\dimexpr\f@size pt\relax}%
  \newcommand*\lineheight[1]{\fontsize{\fsize}{#1\fsize}\selectfont}%
  \ifx\svgwidth\undefined%
    \setlength{\unitlength}{269bp}%
    \ifx\svgscale\undefined%
      \relax%
    \else%
      \setlength{\unitlength}{\unitlength * \real{\svgscale}}%
    \fi%
  \else%
    \setlength{\unitlength}{\svgwidth}%
  \fi%
  \global\let\svgwidth\undefined%
  \global\let\svgscale\undefined%
  \makeatother%
  \begin{picture}(1,0.3477868)%
    \lineheight{1}%
    \setlength\tabcolsep{0pt}%
    \put(0,0){\includegraphics[width=\unitlength,page=1]{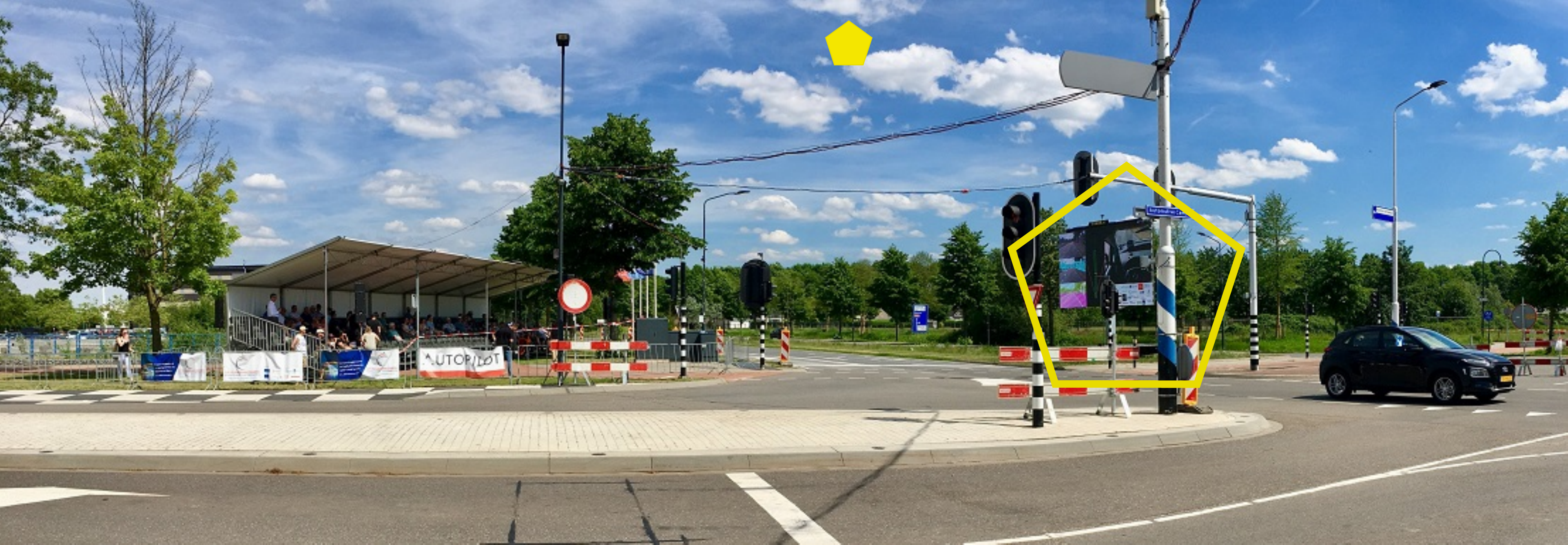}}%
    \put(0.380194927,0.3101328759){\color[rgb]{0,0,0}\makebox(0,0)[lt]{\lineheight{1.25}\smash{\begin{tabular}[t]{l}\textcolor{white}{projector}\end{tabular}}}}%
  \end{picture}%
\endgroup%
}
    
    \vspace*{0.1cm}
    
  \caption{Top left: Outside cameras mounted on car prototype (three of four cameras shown). Top right: The HMI consists of clusters showing the recommended velocity and LED stripes signaling the path choices (left, right or straight). Bottom: Test area with projector visualizing the risk graph.}
  \label{fig:vidas}
\end{figure}

This section presents quantitative experiments for the system combination of RM with R-LDM.
A demonstration took place within the frame of the ITS European Congress 2019 at Helmond, the Netherlands. The system was run live in front of scientific and non-scientific audiences. The main focus lied hereby on the resulting HMI that has been shown and served to warn the driver. The test car and HMI is summarized in Fig. \ref{fig:vidas}. Concretely, by developing an integrated HMI, the suggestions of the systems were rendered transparent.

\subsection{Test Setup}

The used car prototype is called Carlota and was provided by the Spanish research institute Vicomtech. Its sensor setup consists of a mid-range GNSS\footnote{The product model is given on the website https://www.u-blox.com/de/
product/evk-8evk-m8.} 
and cameras. Here, the cameras are four Sekonix
\footnote{See the cameras on http://sekolab.com/products/camera.} 
cameras, which are combined with two Nvidia GPUs\footnote{See https://www.nvidia.com/en-us/geforce/graphics-cards/rtx-2080-ti.} 
and one Intel CPU. For this reason, Carlota can be seen as a low-cost test solution. 

In what follows, we analyze and evaluate the given HMI concept, which is built on three components: a risk graph (top left of Fig. \ref{fig:single_costs}), a velocity scale (Fig. \ref{fig:path_blending}, top right) and lane change recommendations (Fig. \ref{fig:path_blending}, bottom right). 
While demonstrating the system online, the velocity scale was displayed on an instrument cluster, inside the vehicle. For the lane change recommendation, LED stripes were used that were attached to the windshield and are visually conveying suggested driving directions. Additionally, both elements, together with the risk graph, have been presented on a projector to the audience. 

In total, the R-LDM serves to provide environment data, i.e., the paths for the risk calculation. We performed situations with three test drivers, maneuvering Carlota and two further vehicles. In each of those runs, RM supports to execute a safe lane change. The end of the road, indicated by cones, forces a lane change of the ego car. In this context, the speed ranges for the tests lied in between $\unit[20]{km/h}$ and $\unit[40]{km/h}$ with a closed public highway stretch of $\unit[1.5]{km}$ length. Our test area is a two-lane road, with both lanes in the same direction and without oncoming traffic. Data recordings from the demonstration were the basis for the evaluation. 

We now characterize in Section \ref{subsec:integration} benefits that arise from utilizing the R-LDM as a knowledge hub for support systems. Afterwards, within Section \ref{sec:results}, the results of the RM are described and discussed in regard of analyzing the lane change situations.  

\subsection{Sensor Data Fusion}
\label{subsec:integration}
Required inputs for RM are the ego states and its map relation. In a first step, we obtain a (visually) lane-matched ego position from a localization module. The ego vehicle is thus projected onto the lane center, determined by the localization module (i.e., the current ego lane). The R-LDM can now provide driving paths. As a second input of the demonstrated system, we need measured positions and velocities of other cars, see Section \ref{subsec:sensorproc}. 
In this pre-processing step, the four cameras (i.e., left, right, front and back) are used to detect surrounding obstacles in a 360$^{\circ}$ view. Specifically, detected bounding boxes of a YOLOv3 neural network \cite{redmon2018} are projected from each camera image into a joint 3D world. 

In the end, we fuse the pre-processed sensor signals with map data from the R-LDM, the graph-based environment representation. This can give us a driving situation. At every timestep, namely, $N_t\hspace{-0.08cm}=\hspace{-0.07cm}21\hspace{-0.04cm}$ ego trajectories are sampled and predicted with RM ($10$ acceleration and $10$ braking ramps with different acceleration/braking strength, and $1$ constant velocity) to retrieve a situation. Here, sampling is done for the current and parallel lane paths (i.e., $M_p\hspace{-0.06cm}=\hspace{-0.06cm}2$). 

With the middleware RTMaps, we may integrate the software components of R-LDM and RM in a single system. RTMaps reads the sensor inputs and writes final HMI outputs. The system could run with a frequency of around $f\hspace{-0.06cm}=\hspace{-0.06cm}\unit[10]{Hz}$. Our RM can then output a driving path and a velocity that promises safe behavior. As mentioned, the outputs are visualized on the final HMI for driver warning.    

\vspace{-0.15cm}
\subsection{Results}
\label{sec:results}

In the tests, the ego car drives on an ending lane and must take the gap between two vehicles. As we know, the driver may switch to its neighboring lane before or after the passing car (refer to Fig. \ref{fig:demo_scenario}). We will analyze examples of gap and no-gap situations below. These lane changes simultaneously include longitudinal and lateral spatial risks, which make them complex maneuvers. 

\subsubsection{Gap Scenario}
\label{subsec:gap}

\begin{figure}[t!]
  \centering
  \vspace*{-0.035cm}
  \resizebox{1.0\linewidth}{!}{%% Creator: Inkscape inkscape 0.92.4, www.inkscape.org
%% PDF/EPS/PS + LaTeX output extension by Johan Engelen, 2010
%% Accompanies image file 'gap_scenario_shot1.pdf' (pdf, eps, ps)
%%
%% To include the image in your LaTeX document, write
%%   \input{<filename>.pdf_tex}
%%  instead of
%%   \includegraphics{<filename>.pdf}
%% To scale the image, write
%%   \def\svgwidth{<desired width>}
%%   \input{<filename>.pdf_tex}
%%  instead of
%%   \includegraphics[width=<desired width>]{<filename>.pdf}
%%
%% Images with a different path to the parent latex file can
%% be accessed with the `import' package (which may need to be
%% installed) using
%%   \usepackage{import}
%% in the preamble, and then including the image with
%%   \import{<path to file>}{<filename>.pdf_tex}
%% Alternatively, one can specify
%%   \graphicspath{{<path to file>/}}
%% 
%% For more information, please see info/svg-inkscape on CTAN:
%%   http://tug.ctan.org/tex-archive/info/svg-inkscape
%%
\begingroup%
  \makeatletter%
  \providecommand\color[2][]{%
    \errmessage{(Inkscape) Color is used for the text in Inkscape, but the package 'color.sty' is not loaded}%
    \renewcommand\color[2][]{}%
  }%
  \providecommand\transparent[1]{%
    \errmessage{(Inkscape) Transparency is used (non-zero) for the text in Inkscape, but the package 'transparent.sty' is not loaded}%
    \renewcommand\transparent[1]{}%
  }%
  \providecommand\rotatebox[2]{#2}%
  \newcommand*\fsize{\dimexpr\f@size pt\relax}%
  \newcommand*\lineheight[1]{\fontsize{\fsize}{#1\fsize}\selectfont}%
  \ifx\svgwidth\undefined%
    \setlength{\unitlength}{337.59468812bp}%
    \ifx\svgscale\undefined%
      \relax%
    \else%
      \setlength{\unitlength}{\unitlength * \real{\svgscale}}%
    \fi%
  \else%
    \setlength{\unitlength}{\svgwidth}%
  \fi%
  \global\let\svgwidth\undefined%
  \global\let\svgscale\undefined%
  \makeatother%
  \begin{picture}(1,0.21031625)%
    \lineheight{1}%
    \setlength\tabcolsep{0pt}%
    \put(0,0){\includegraphics[width=\unitlength,page=1]{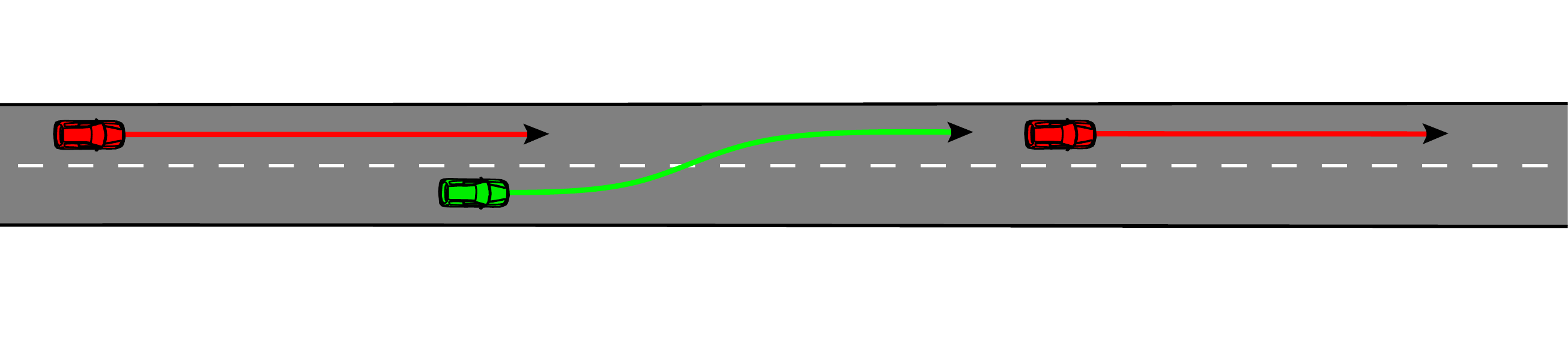}}%
    \put(0.62386105,0.17625989){\color[rgb]{0,0,0}\makebox(0,0)[lt]{\lineheight{1.25}\smash{\begin{tabular}[t]{l}$d_2 \hspace{-0.03cm} = \hspace{-0.03cm} \unit[22]{m}$\end{tabular}}}}%
    \put(0.02184035,0.17605989){\color[rgb]{0,0,0}\makebox(0,0)[lt]{\lineheight{1.25}\smash{\begin{tabular}[t]{l}$d_1 \hspace{-0.03cm} = \hspace{-0.03cm} \unit[19]{m}$\end{tabular}}}}%
    \put(0.25514661,0.03725136){\color[rgb]{0,0,0}\makebox(0,0)[lt]{\lineheight{1.25}\smash{\begin{tabular}[t]{l}$v_0 \hspace{-0.03cm} = \hspace{-0.03cm} \unit[8]{m/\text{sec}}$\end{tabular}}}}%
    \put(-0.29038696,-0.29862192){\color[rgb]{0,0,0}\makebox(0,0)[lt]{\begin{minipage}{0.08095144\unitlength}\raggedright \end{minipage}}}%
    \put(-1.24955896,-0.24760213){\color[rgb]{0,0,0}\makebox(0,0)[lt]{\begin{minipage}{0.18979361\unitlength}\raggedright \end{minipage}}}%
  \end{picture}%
\endgroup%
}
  
  \vspace{-0.094cm}
  
  \resizebox{0.85\linewidth}{!}{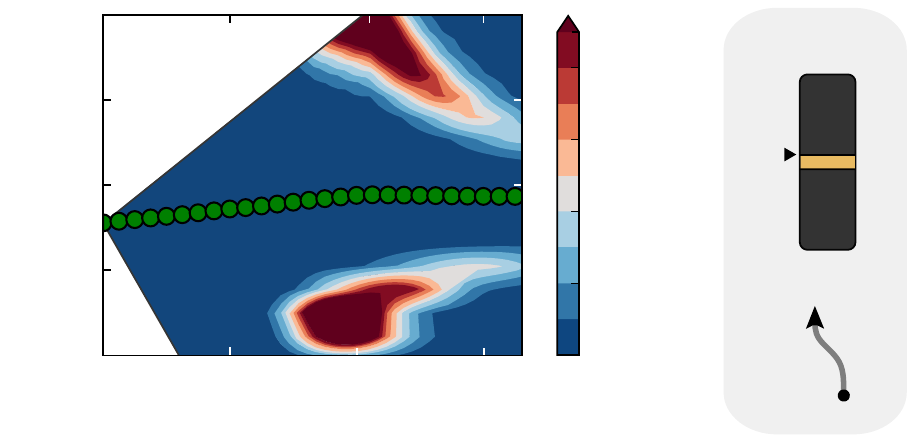\hspace{0.8cm}}
  
  \vspace{0.37cm}
  
  \resizebox{1.0\linewidth}{!}{%% Creator: Inkscape inkscape 0.92.4, www.inkscape.org
%% PDF/EPS/PS + LaTeX output extension by Johan Engelen, 2010
%% Accompanies image file 'gap_scenario_shot2.pdf' (pdf, eps, ps)
%%
%% To include the image in your LaTeX document, write
%%   \input{<filename>.pdf_tex}
%%  instead of
%%   \includegraphics{<filename>.pdf}
%% To scale the image, write
%%   \def\svgwidth{<desired width>}
%%   \input{<filename>.pdf_tex}
%%  instead of
%%   \includegraphics[width=<desired width>]{<filename>.pdf}
%%
%% Images with a different path to the parent latex file can
%% be accessed with the `import' package (which may need to be
%% installed) using
%%   \usepackage{import}
%% in the preamble, and then including the image with
%%   \import{<path to file>}{<filename>.pdf_tex}
%% Alternatively, one can specify
%%   \graphicspath{{<path to file>/}}
%% 
%% For more information, please see info/svg-inkscape on CTAN:
%%   http://tug.ctan.org/tex-archive/info/svg-inkscape
%%
\begingroup%
  \makeatletter%
  \providecommand\color[2][]{%
    \errmessage{(Inkscape) Color is used for the text in Inkscape, but the package 'color.sty' is not loaded}%
    \renewcommand\color[2][]{}%
  }%
  \providecommand\transparent[1]{%
    \errmessage{(Inkscape) Transparency is used (non-zero) for the text in Inkscape, but the package 'transparent.sty' is not loaded}%
    \renewcommand\transparent[1]{}%
  }%
  \providecommand\rotatebox[2]{#2}%
  \newcommand*\fsize{\dimexpr\f@size pt\relax}%
  \newcommand*\lineheight[1]{\fontsize{\fsize}{#1\fsize}\selectfont}%
  \ifx\svgwidth\undefined%
    \setlength{\unitlength}{337.59468812bp}%
    \ifx\svgscale\undefined%
      \relax%
    \else%
      \setlength{\unitlength}{\unitlength * \real{\svgscale}}%
    \fi%
  \else%
    \setlength{\unitlength}{\svgwidth}%
  \fi%
  \global\let\svgwidth\undefined%
  \global\let\svgscale\undefined%
  \makeatother%
  \begin{picture}(1,0.21031625)%
    \lineheight{1}%
    \setlength\tabcolsep{0pt}%
    \put(0,0){\includegraphics[width=\unitlength,page=1]{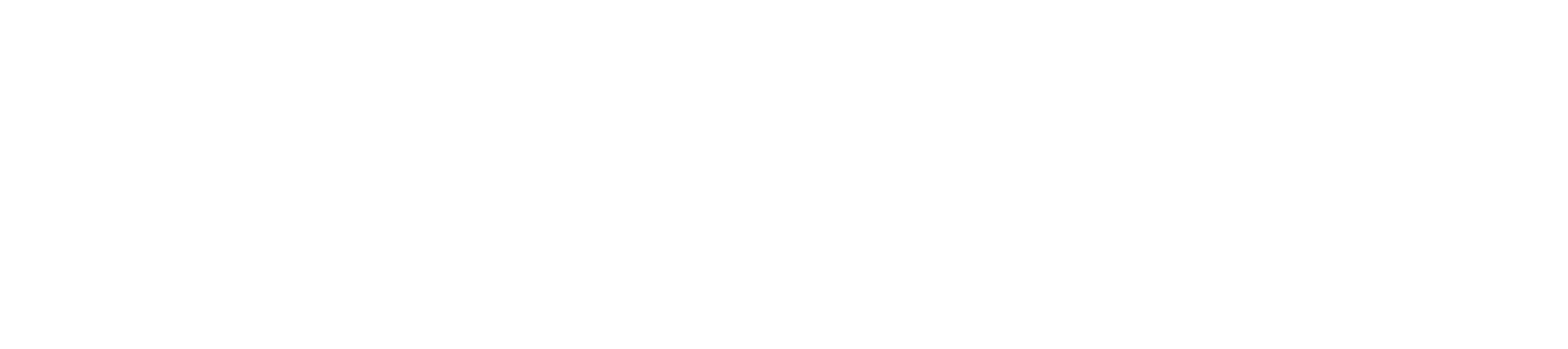}}%
    \put(0.07553538,0.17905989){\color[rgb]{0,0,0}\makebox(0,0)[lt]{\lineheight{1.25}\smash{\begin{tabular}[t]{l}$d_1 \hspace{-0.03cm} = \hspace{-0.03cm} \unit[16]{m}$\end{tabular}}}}%
    \put(0,0){\includegraphics[width=\unitlength,page=2]{gap_scenario_shot2.pdf}}%
    \put(0.67086105,0.17905989){\color[rgb]{0,0,0}\makebox(0,0)[lt]{\lineheight{1.25}\smash{\begin{tabular}[t]{l}$d_2 \hspace{-0.03cm} = \hspace{-0.03cm} \unit[22]{m}$\end{tabular}}}}%
    \put(0.30021662,0.03664361){\color[rgb]{0,0,0}\makebox(0,0)[lt]{\lineheight{1.25}\smash{\begin{tabular}[t]{l}$v_0 \hspace{-0.03cm} = \hspace{-0.03cm} \unit[7]{m/\text{sec}}$\end{tabular}}}}%
    \put(-0.29025407,-0.02690786){\color[rgb]{0,0,0}\makebox(0,0)[lt]{\begin{minipage}{0.08068475\unitlength}\raggedright \end{minipage}}}%
    \put(-1.2462661,0.02394384){\color[rgb]{0,0,0}\makebox(0,0)[lt]{\begin{minipage}{0.18916834\unitlength}\raggedright \end{minipage}}}%
  \end{picture}%
\endgroup%
}
  
  \vspace{-0.075cm}
  
  \resizebox{0.85\linewidth}{!}{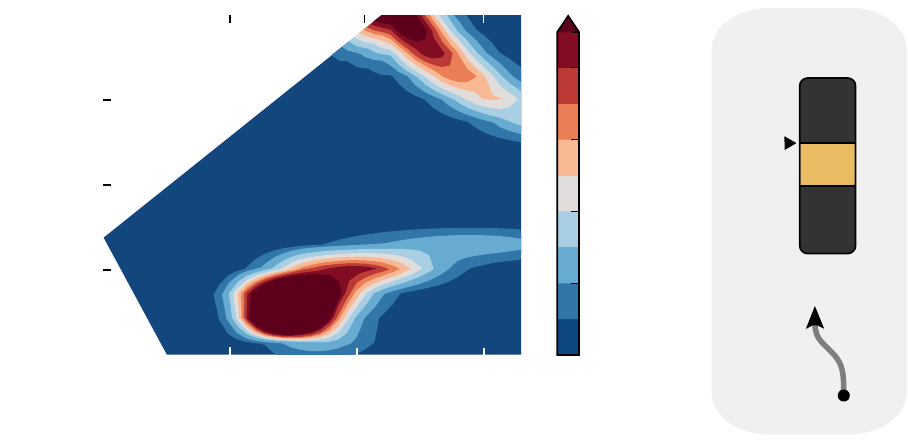\hspace{0.8cm}}
  
  \vspace{0.31cm}
  
  \resizebox{1.0\linewidth}{!}{%% Creator: Inkscape inkscape 0.92.4, www.inkscape.org
%% PDF/EPS/PS + LaTeX output extension by Johan Engelen, 2010
%% Accompanies image file 'gap_scenario_shot2.pdf' (pdf, eps, ps)
%%
%% To include the image in your LaTeX document, write
%%   \input{<filename>.pdf_tex}
%%  instead of
%%   \includegraphics{<filename>.pdf}
%% To scale the image, write
%%   \def\svgwidth{<desired width>}
%%   \input{<filename>.pdf_tex}
%%  instead of
%%   \includegraphics[width=<desired width>]{<filename>.pdf}
%%
%% Images with a different path to the parent latex file can
%% be accessed with the `import' package (which may need to be
%% installed) using
%%   \usepackage{import}
%% in the preamble, and then including the image with
%%   \import{<path to file>}{<filename>.pdf_tex}
%% Alternatively, one can specify
%%   \graphicspath{{<path to file>/}}
%% 
%% For more information, please see info/svg-inkscape on CTAN:
%%   http://tug.ctan.org/tex-archive/info/svg-inkscape
%%
\begingroup%
  \makeatletter%
  \providecommand\color[2][]{%
    \errmessage{(Inkscape) Color is used for the text in Inkscape, but the package 'color.sty' is not loaded}%
    \renewcommand\color[2][]{}%
  }%
  \providecommand\transparent[1]{%
    \errmessage{(Inkscape) Transparency is used (non-zero) for the text in Inkscape, but the package 'transparent.sty' is not loaded}%
    \renewcommand\transparent[1]{}%
  }%
  \providecommand\rotatebox[2]{#2}%
  \newcommand*\fsize{\dimexpr\f@size pt\relax}%
  \newcommand*\lineheight[1]{\fontsize{\fsize}{#1\fsize}\selectfont}%
  \ifx\svgwidth\undefined%
    \setlength{\unitlength}{337.59468812bp}%
    \ifx\svgscale\undefined%
      \relax%
    \else%
      \setlength{\unitlength}{\unitlength * \real{\svgscale}}%
    \fi%
  \else%
    \setlength{\unitlength}{\svgwidth}%
  \fi%
  \global\let\svgwidth\undefined%
  \global\let\svgscale\undefined%
  \makeatother%
  \begin{picture}(1,0.21031625)%
    \lineheight{1}%
    \setlength\tabcolsep{0pt}%
    \put(0,0){\includegraphics[width=\unitlength,page=1]{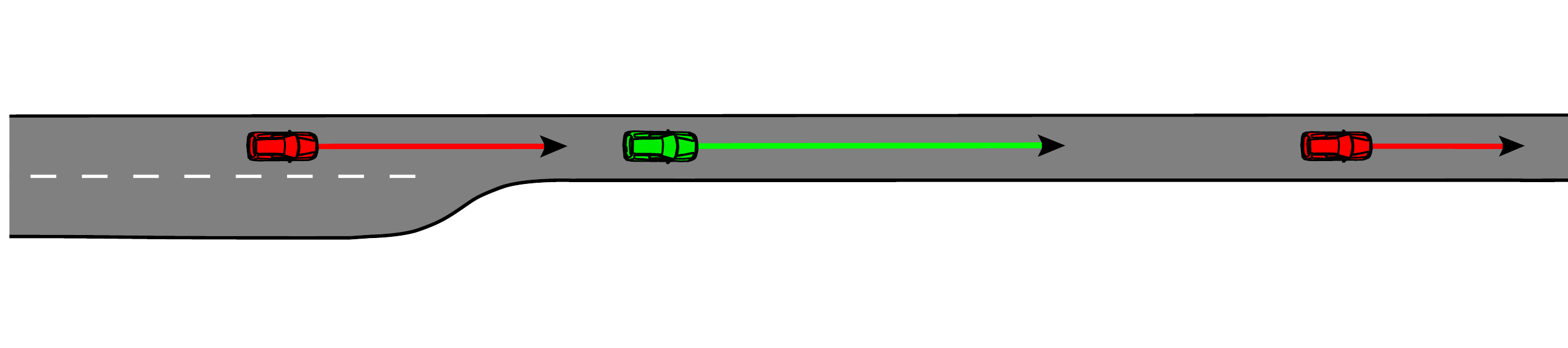}}%
    \put(0.17653538,0.16975989){\color[rgb]{0,0,0}\makebox(0,0)[lt]{\lineheight{1.25}\smash{\begin{tabular}[t]{l}$d_1 \hspace{-0.03cm} = \hspace{-0.03cm} \unit[16]{m}$\end{tabular}}}}%
    \put(0.76886105,0.16975989){\color[rgb]{0,0,0}\makebox(0,0)[lt]{\lineheight{1.25}\smash{\begin{tabular}[t]{l}$d_2 \hspace{-0.03cm} = \hspace{-0.03cm} \unit[25]{m}$\end{tabular}}}}%
    \put(0.40021662,0.06534361){\color[rgb]{0,0,0}\makebox(0,0)[lt]{\lineheight{1.25}\smash{\begin{tabular}[t]{l}$v_0 \hspace{-0.03cm} = \hspace{-0.03cm} \unit[7]{m/\text{sec}}$\end{tabular}}}}%
    \put(-0.29025407,-0.02690786){\color[rgb]{0,0,0}\makebox(0,0)[lt]{\begin{minipage}{0.08068475\unitlength}\raggedright \end{minipage}}}%
    \put(-1.2462661,0.02394384){\color[rgb]{0,0,0}\makebox(0,0)[lt]{\begin{minipage}{0.18916834\unitlength}\raggedright \end{minipage}}}%
  \end{picture}%
\endgroup%
}
  
  \vspace{-0.21cm}
  
  \resizebox{0.85\linewidth}{!}{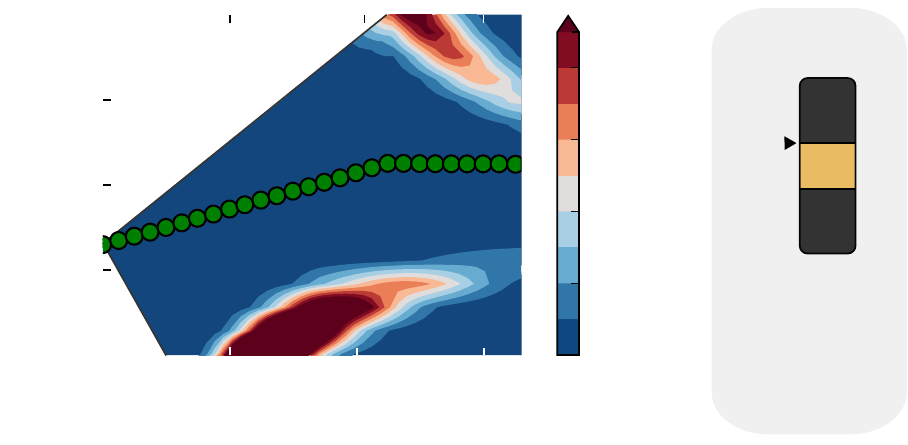\hspace{0.8cm}}
  \vspace{0.058cm}
  \caption{Behavior of RM and R-LDM in a gap scenario. Given are three snapshots from real-world recordings. The system signals to directly perform a lane change due to sufficient space between the neighboring cars. Top: Road layout and predictive situation, Bottom: HMI with risk visualizations.}   
  \label{fig:gap_scenario}
\end{figure}

For the scenario, recordings have been replayed with three vehicles and, subsequently, we predict their trajectories. Fig. \ref{fig:gap_scenario} pictures the different situation snapshots with the ego velocity $v_0$ as well as distances $d_1$ and $d_2$ to the two other cars on the neighboring lane. Herein, the planned ego trajectory is colored green, with the predicted trajectories of surrounding vehicles colored red. On average, the other cars drive in the complete stream with constant velocity.\footnote{Note that the other trajectories' length relates to a constant velocity, while the ego trajectories incorporate acceleration or deceleration.} 

A planned lane change is realized by blending the current path with the parallel path at a start time $t_{\text{start}}\hspace{-0.05cm}=\hspace{-0.05cm}\unit[1]{\text{sec}}$ with a duration of $\unit[3]{\text{sec}}$ (for details, see Section \ref{subsec:lat}). This aligns with usual lane change durations, e.g., according to \cite{toledo2007}. 
For the visualizations, we choose a prediction horizon $s_h$ of $\unit[6]{\text{sec}}$ but the actual risk is evaluated for $s_h\hspace{-0.06cm}=\hspace{-0.06cm}\unit[12]{\text{sec}}$. The car signals and possible paths are updated in the R-LDM on demand. 

Fig. \ref{fig:gap_scenario} also illustrates the risk graph. The sampled ego trajectories are plotted as curves of velocity over the future time and the sum of of collision and curve rates with $\tau^{-1}_{\text{crit}}$ are further visualized. The blue areas in this graph represent low probabilities between $[\unit[0]{\%/\text{sec}},\unit[0.5]{\%/\text{sec}}]$ and red areas are high values with $[\unit[0.5]{\%/\text{sec}}$ and also values $>\hspace{-0.085cm}\unit[1]{\%/\text{sec}}]$. Finally, the single chosen trajectory is highlighted with green points. RM always tries to find a trajectory that bypasses the red hot spots. 

In this scenario, RM successfully judges the gap as sufficiently large. Changing the lane with moderate acceleration from $\unit[7]{m/\text{sec}}$ to $v_{\text{tar}}\hspace{-0.04cm}=\hspace{-0.04cm}\unit[11]{m/\text{sec}}$ presents the optimal maneuver. After the lane change, the message on the HMI changes from "go left" to "go straight" (i.e., the target path $\textbf{p}_{\text{tar}}$). This example emphasizes the possible proactive support that can be provided by RM. 

\begin{figure}[t!]
  \centering
  \vspace*{-0.05cm}
  \resizebox{1.0\linewidth}{!}{%% Creator: Inkscape inkscape 0.92.4, www.inkscape.org
%% PDF/EPS/PS + LaTeX output extension by Johan Engelen, 2010
%% Accompanies image file 'gap_scenario_shot2.pdf' (pdf, eps, ps)
%%
%% To include the image in your LaTeX document, write
%%   \input{<filename>.pdf_tex}
%%  instead of
%%   \includegraphics{<filename>.pdf}
%% To scale the image, write
%%   \def\svgwidth{<desired width>}
%%   \input{<filename>.pdf_tex}
%%  instead of
%%   \includegraphics[width=<desired width>]{<filename>.pdf}
%%
%% Images with a different path to the parent latex file can
%% be accessed with the `import' package (which may need to be
%% installed) using
%%   \usepackage{import}
%% in the preamble, and then including the image with
%%   \import{<path to file>}{<filename>.pdf_tex}
%% Alternatively, one can specify
%%   \graphicspath{{<path to file>/}}
%% 
%% For more information, please see info/svg-inkscape on CTAN:
%%   http://tug.ctan.org/tex-archive/info/svg-inkscape
%%
\begingroup%
  \makeatletter%
  \providecommand\color[2][]{%
    \errmessage{(Inkscape) Color is used for the text in Inkscape, but the package 'color.sty' is not loaded}%
    \renewcommand\color[2][]{}%
  }%
  \providecommand\transparent[1]{%
    \errmessage{(Inkscape) Transparency is used (non-zero) for the text in Inkscape, but the package 'transparent.sty' is not loaded}%
    \renewcommand\transparent[1]{}%
  }%
  \providecommand\rotatebox[2]{#2}%
  \newcommand*\fsize{\dimexpr\f@size pt\relax}%
  \newcommand*\lineheight[1]{\fontsize{\fsize}{#1\fsize}\selectfont}%
  \ifx\svgwidth\undefined%
    \setlength{\unitlength}{337.59468812bp}%
    \ifx\svgscale\undefined%
      \relax%
    \else%
      \setlength{\unitlength}{\unitlength * \real{\svgscale}}%
    \fi%
  \else%
    \setlength{\unitlength}{\svgwidth}%
  \fi%
  \global\let\svgwidth\undefined%
  \global\let\svgscale\undefined%
  \makeatother%
  \begin{picture}(1,0.21031625)%
    \lineheight{1}%
    \setlength\tabcolsep{0pt}%
    \put(0,0){\includegraphics[width=\unitlength,page=1]{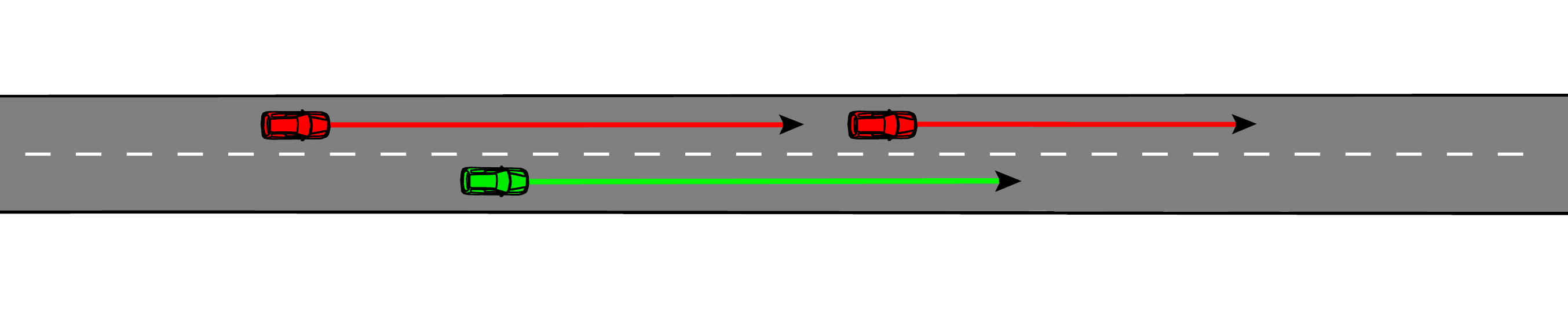}}%
    \put(0.14453538,0.17105989){\color[rgb]{0,0,0}\makebox(0,0)[lt]{\lineheight{1.25}\smash{\begin{tabular}[t]{l}$d_1 \hspace{-0.03cm} = \hspace{-0.03cm} \unit[8]{m}$\end{tabular}}}}%
    \put(0.50086105,0.17105989){\color[rgb]{0,0,0}\makebox(0,0)[lt]{\lineheight{1.25}\smash{\begin{tabular}[t]{l}$d_2 \hspace{-0.03cm} = \hspace{-0.03cm} \unit[15]{m}$\end{tabular}}}}%
    \put(0.25521662,0.03744361){\color[rgb]{0,0,0}\makebox(0,0)[lt]{\lineheight{1.25}\smash{\begin{tabular}[t]{l}$v_0 \hspace{-0.03cm} = \hspace{-0.03cm} \unit[9]{m/\text{sec}}$\end{tabular}}}}%
    \put(-0.29025407,-0.02690786){\color[rgb]{0,0,0}\makebox(0,0)[lt]{\begin{minipage}{0.08068475\unitlength}\raggedright \end{minipage}}}%
    \put(-1.2462661,0.02394384){\color[rgb]{0,0,0}\makebox(0,0)[lt]{\begin{minipage}{0.18916834\unitlength}\raggedright \end{minipage}}}%
  \end{picture}%
\endgroup%
}
  
  \vspace{-0.04cm}
  
  \resizebox{0.85\linewidth}{!}{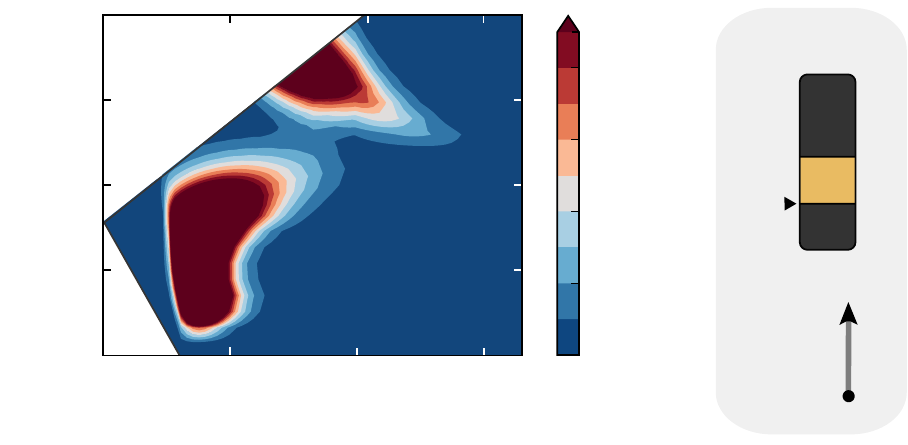\hspace{0.8cm}}
  
  \vspace{0.23cm}
  
  \resizebox{1.0\linewidth}{!}{%% Creator: Inkscape inkscape 0.92.4, www.inkscape.org
%% PDF/EPS/PS + LaTeX output extension by Johan Engelen, 2010
%% Accompanies image file 'gap_scenario_shot2.pdf' (pdf, eps, ps)
%%
%% To include the image in your LaTeX document, write
%%   \input{<filename>.pdf_tex}
%%  instead of
%%   \includegraphics{<filename>.pdf}
%% To scale the image, write
%%   \def\svgwidth{<desired width>}
%%   \input{<filename>.pdf_tex}
%%  instead of
%%   \includegraphics[width=<desired width>]{<filename>.pdf}
%%
%% Images with a different path to the parent latex file can
%% be accessed with the `import' package (which may need to be
%% installed) using
%%   \usepackage{import}
%% in the preamble, and then including the image with
%%   \import{<path to file>}{<filename>.pdf_tex}
%% Alternatively, one can specify
%%   \graphicspath{{<path to file>/}}
%% 
%% For more information, please see info/svg-inkscape on CTAN:
%%   http://tug.ctan.org/tex-archive/info/svg-inkscape
%%
\begingroup%
  \makeatletter%
  \providecommand\color[2][]{%
    \errmessage{(Inkscape) Color is used for the text in Inkscape, but the package 'color.sty' is not loaded}%
    \renewcommand\color[2][]{}%
  }%
  \providecommand\transparent[1]{%
    \errmessage{(Inkscape) Transparency is used (non-zero) for the text in Inkscape, but the package 'transparent.sty' is not loaded}%
    \renewcommand\transparent[1]{}%
  }%
  \providecommand\rotatebox[2]{#2}%
  \newcommand*\fsize{\dimexpr\f@size pt\relax}%
  \newcommand*\lineheight[1]{\fontsize{\fsize}{#1\fsize}\selectfont}%
  \ifx\svgwidth\undefined%
    \setlength{\unitlength}{337.59468812bp}%
    \ifx\svgscale\undefined%
      \relax%
    \else%
      \setlength{\unitlength}{\unitlength * \real{\svgscale}}%
    \fi%
  \else%
    \setlength{\unitlength}{\svgwidth}%
  \fi%
  \global\let\svgwidth\undefined%
  \global\let\svgscale\undefined%
  \makeatother%
  \begin{picture}(1,0.21031625)%
    \lineheight{1}%
    \setlength\tabcolsep{0pt}%
    \put(0,0){\includegraphics[width=\unitlength,page=1]{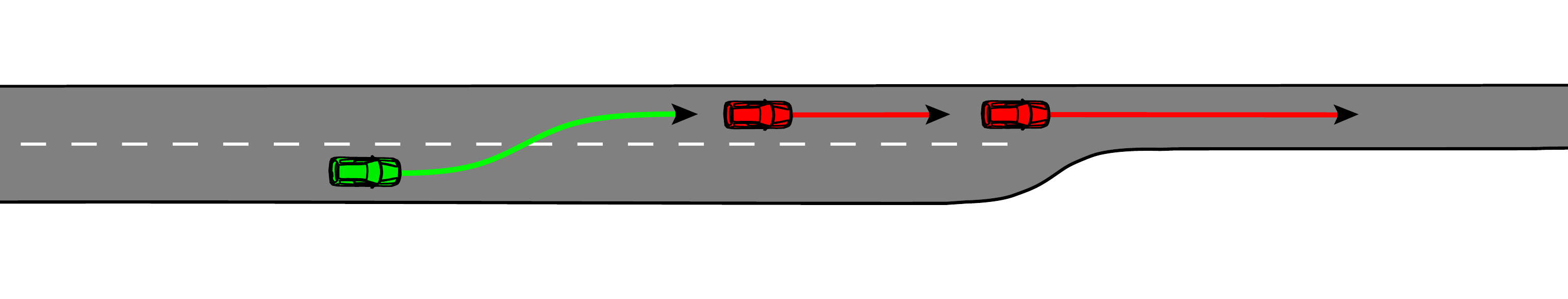}}%
    \put(0.42653538,0.15415989){\color[rgb]{0,0,0}\makebox(0,0)[lt]{\lineheight{1.25}\smash{\begin{tabular}[t]{l}$d_1 \hspace{-0.03cm} = \hspace{-0.03cm} \unit[12]{m}$\end{tabular}}}}%
    \put(0.59086105,0.15515989){\color[rgb]{0,0,0}\makebox(0,0)[lt]{\lineheight{1.25}\smash{\begin{tabular}[t]{l}$d_2 \hspace{-0.03cm} = \hspace{-0.03cm} \unit[20]{m}$\end{tabular}}}}%
    \put(0.16021662,0.01944361){\color[rgb]{0,0,0}\makebox(0,0)[lt]{\lineheight{1.25}\smash{\begin{tabular}[t]{l}$v_0 \hspace{-0.03cm} = \hspace{-0.03cm} \unit[5]{m/\text{sec}}$\end{tabular}}}}%
    \put(-0.29025407,-0.02690786){\color[rgb]{0,0,0}\makebox(0,0)[lt]{\begin{minipage}{0.08068475\unitlength}\raggedright \end{minipage}}}%
    \put(-1.2462661,0.02394384){\color[rgb]{0,0,0}\makebox(0,0)[lt]{\begin{minipage}{0.18916834\unitlength}\raggedright \end{minipage}}}%
  \end{picture}%
\endgroup%
}
  
  \vspace{0.075cm}
  
  \resizebox{0.85\linewidth}{!}{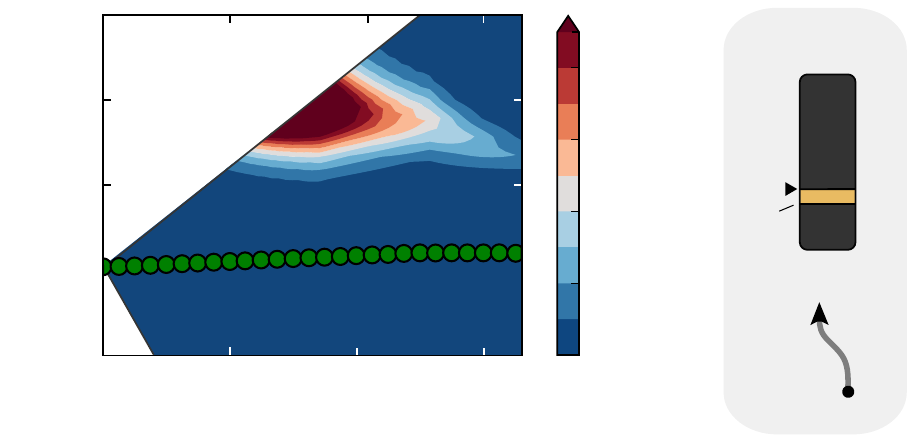\hspace{0.8cm}}
  
  \vspace{0.37cm}
  
  \resizebox{1.0\linewidth}{!}{%% Creator: Inkscape inkscape 0.92.4, www.inkscape.org
%% PDF/EPS/PS + LaTeX output extension by Johan Engelen, 2010
%% Accompanies image file 'gap_scenario_shot2.pdf' (pdf, eps, ps)
%%
%% To include the image in your LaTeX document, write
%%   \input{<filename>.pdf_tex}
%%  instead of
%%   \includegraphics{<filename>.pdf}
%% To scale the image, write
%%   \def\svgwidth{<desired width>}
%%   \input{<filename>.pdf_tex}
%%  instead of
%%   \includegraphics[width=<desired width>]{<filename>.pdf}
%%
%% Images with a different path to the parent latex file can
%% be accessed with the `import' package (which may need to be
%% installed) using
%%   \usepackage{import}
%% in the preamble, and then including the image with
%%   \import{<path to file>}{<filename>.pdf_tex}
%% Alternatively, one can specify
%%   \graphicspath{{<path to file>/}}
%% 
%% For more information, please see info/svg-inkscape on CTAN:
%%   http://tug.ctan.org/tex-archive/info/svg-inkscape
%%
\begingroup%
  \makeatletter%
  \providecommand\color[2][]{%
    \errmessage{(Inkscape) Color is used for the text in Inkscape, but the package 'color.sty' is not loaded}%
    \renewcommand\color[2][]{}%
  }%
  \providecommand\transparent[1]{%
    \errmessage{(Inkscape) Transparency is used (non-zero) for the text in Inkscape, but the package 'transparent.sty' is not loaded}%
    \renewcommand\transparent[1]{}%
  }%
  \providecommand\rotatebox[2]{#2}%
  \newcommand*\fsize{\dimexpr\f@size pt\relax}%
  \newcommand*\lineheight[1]{\fontsize{\fsize}{#1\fsize}\selectfont}%
  \ifx\svgwidth\undefined%
    \setlength{\unitlength}{337.59468812bp}%
    \ifx\svgscale\undefined%
      \relax%
    \else%
      \setlength{\unitlength}{\unitlength * \real{\svgscale}}%
    \fi%
  \else%
    \setlength{\unitlength}{\svgwidth}%
  \fi%
  \global\let\svgwidth\undefined%
  \global\let\svgscale\undefined%
  \makeatother%
  \begin{picture}(1,0.21031625)%
    \lineheight{1}%
    \setlength\tabcolsep{0pt}%
    \put(0,0){\includegraphics[width=\unitlength,page=1]{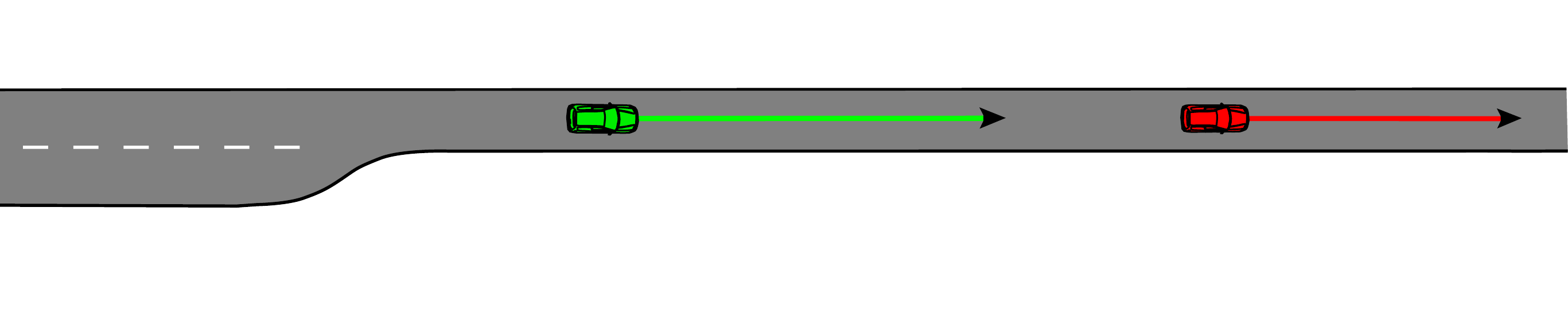}}%
    \put(0.71886105,0.17445989){\color[rgb]{0,0,0}\makebox(0,0)[lt]{\lineheight{1.25}\smash{\begin{tabular}[t]{l}$d_1 \hspace{-0.03cm} = \hspace{-0.03cm} \unit[21]{m}$\end{tabular}}}}%
    \put(0.30521662,0.07434361){\color[rgb]{0,0,0}\makebox(0,0)[lt]{\lineheight{1.25}\smash{\begin{tabular}[t]{l}$v_0 \hspace{-0.03cm} = \hspace{-0.03cm} \unit[6]{m/\text{sec}}$\end{tabular}}}}%
    \put(-0.29025407,-0.02690786){\color[rgb]{0,0,0}\makebox(0,0)[lt]{\begin{minipage}{0.08068475\unitlength}\raggedright \end{minipage}}}%
    \put(-1.2462661,0.02394384){\color[rgb]{0,0,0}\makebox(0,0)[lt]{\begin{minipage}{0.18916834\unitlength}\raggedright \end{minipage}}}%
  \end{picture}%
\endgroup%
}
  
  \vspace{-0.30cm}
  
  \resizebox{0.85\linewidth}{!}{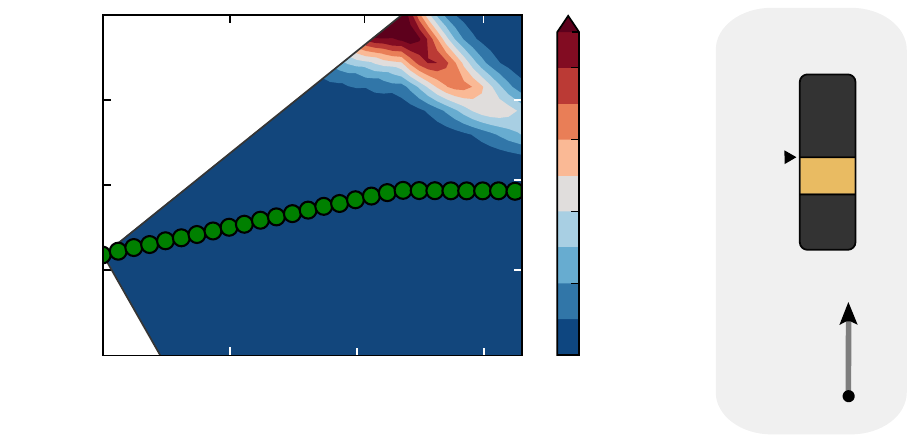\hspace{0.8cm}}
  \vspace{0.032cm}
  \caption{Behavior of RM and R-LDM in a no-gap scenario. A forced lane change with an acceleration is advised only after the preceding car passes. This represents a safe motion and the driver follows the warning.} 
  \label{fig:no_gap_scenario}
\end{figure} 

\subsubsection{No-Gap Scenario}
\label{subsec:nogap}
In the previous gap scenario, the system recommended to either drive with constant velocity or to accelerate for performing the lane change. The front and back car were determining the gap. In this second scenario, the gap is too small for safely conducting a lane change. Only after the two other vehicles have passed and drive with appropriate distance to the ego vehicle, RM signalizes that the driver could make a lane change.

Within Fig. \ref{fig:no_gap_scenario}, the preceding car hinders the ego car to go on the target lane. Hereby, we want to analyze more closely the risk graph for the first screenshot of the scenario. If the ego vehicle brakes or drives constantly and makes the lane change, high risks are caused by the preceding vehicle. Due to the close lateral distance, a large risk spot is depicted in the bottom left of the graph. Secondly, if the ego car accelerates and changes lane, there is a high risk of colliding with the front vehicle, indicated by the other red area. The risk graph shows the reasoning behind the system's recommendation output: ``brake'' and ``stay  on the lane''. 

The actual driven speed $v_0$ and target speed $v_{\text{tar}}$ is depicted in the velocity scale. As a reminder, the comfort velocity parameter $v_d$ determines a utility gain and can be chosen by the user. In Fig. \ref{fig:no_gap_scenario}, the system penalizes velocities that deviate from $v_d=\unit[10]{m/\text{sec}}$. To find an ego trajectory in terms of risk versus comfort, we sanction high accelerations as well. RM, therefore, initially recommends to brake until $\unit[4]{m/\text{sec}}$ because of the ending lane. When the preceding vehicle passed the ego vehicle, RM then correctly recommends to drive with constant velocity $\unit[6]{m/\text{sec}}$ and $v_{\text{tar}}$ is lastly accelerating until $\unit[9]{m/\text{sec}}$. The ego driver's speed stays around $v_0=\unit[5]{m/\text{sec}}$ and the lane change suggestion is followed.

\vspace{-0.06cm}
\subsection{Discussion}
\label{subsec:noise}
As we could see, the combination of RM and R-LDM allowed for a reasonable support of the driver in lane change scenarios. In this Section \ref{subsec:noise}, a short discussion concerning the results of the system is given. After showing insights into noise robustness with real-world sensors, the generalization for other driving situations is discussed.

Noise in the vehicles' position and velocity due to the real cameras and GNSS sensors can potentially hinder a faultless operation. In the system, we parametrized the weights in the planning (i.e., for risk, utility and comfort) to achieve robustness against such noise errors. 
As explained in Section \ref{subsec:sensorproc}, positions of all vehicles are projected to the closest path, derived from the R-LDM. With regard to risk computations, the noise did not affect the lane change advices of our experiments. Position noise that is constantly present is merely carried into RM in the future time, where the survival analysis compensates its influence. 

However, the safe speed output from the planner can slightly fluctuate. This is due to the velocity noise in the ego and other vehicles, which grows over the prediction time. Furthermore, a projection on wrong lanes induces the biggest errors. To tackle this noise propagation and the impact of discrete errors, we introduce a hysteresis in RM.
The hysteresis changes the outputs solely when the new selection shows lower risk values for at least $\unit[2]{\text{sec}}$. 
The more RM is robust against this sensor noise, the less it will proactively react on the changes in the environment. In the end, a tradeoff was chosen between stable warnings versus better sensitivity.  

The given experiments show that a map-based behavior support works online and has benefits in terms of its predictability and transparent risk model. The last Section \ref{sec:results} served as a first proof of concept that RM can be applied in real-time on a car platform. The tests were conducted with velocities of $\unit[30]{km/h}$. Nevertheless, other applications on Spanish city streets (i.e., $\unit[50]{km/h}$ limit) and German highway entrances (on average, $\unit[100]{km/h}$) show the same functionalities and beneficial properties. For these test cases, we applied RM on other car platforms using higher-cost GNSS and lidar.  

In addition, in previous scientific work we showed that the same risk-based planner can be successfully applied in simulation to turning at curves (see \cite{puphal2018}), merging at intersections (see also \cite{puphal2018}) or crossing at intersections (see \cite{puphal2019}). For this reason, we consider the planner to be generalizable to various traffic cases in urban or highway driving due to the generic characteristics of the models. The transparent model-based risk approach allows to visualize and understand the reasoning behind its decisions.

\section{Conclusion and Outlook}
\label{sec:outl}

\urlstyle{rm} 
In summary, this paper served to evaluate an application of the Risk Maps (RM) technology on a prototype vehicle in order to show its real-world applicability. In an online demonstration, we presented warning functions for the example of forced lane changes. For this purpose, we extended RM to handle lateral path options.

The system utilizes the Relational Local Dynamic Map (R-LDM) for the alignment, filtering and dynamic layering of car signals with map data. Afterwards, RM are used to predict the current traffic situation into the future and probe trajectories and paths based on driving risks for planning. 
An improved GNSS position hereby allows to position the ego vehicle on the road. RM can handle noise of the camera sensors for object detection by considering uncertainty by design. In addition, the visualization of a risk graph supports the human driver to understand the driving situation. 

Our resulting ADAS was demonstrated on the ITS European Congress 2019 in Eindhoven, the Netherlands. Videos of the results can be found on Youtube: \url{https://www.youtube.com/watch?v=8o3hT3H_gDU}. The real-world evaluation shows that the system is capable of giving reasonable advice on target speeds and lane changes, separated in gap and no-gap situations. Effectively, we may improve the prediction capability for the user with a risk-based planner. RM represents a white-box model, which handles sensor noise and complex predictions. This is promising to increase trust on ADAS.

In the current realization of the system, we assume path blending with fixed parameters of duration time. In a revised solution, the parameters should be optimized \cite{weisswange2019}. Helpful are heuristics to make sure that the lane change is conducted with sufficient space to the preceding and leading vehicle. By using general risks, we believe that RM can implicitly consider dependencies between velocities and paths. This is especially helpful for autonomous driving.

For future research, we likewise envision the use of situation classifiers to improve risk predictions. Potential behavior of other cars need to be explicitly reflected (i.e., lane changes, accelerations or decelerations). Currently, interaction between vehicles is only modeled by a general increase of uncertainty in the trajectories. However, for interaction-intensive situations (e.g., lane changing in dense traffic, abruptly stopping vehicles, suddenly appearing occluded vehicles or similar dynamic conditions), the intention of other cars need to be predicted. Hereby, machine learning methods offer themselves to predict trajectories based on previously measured positions. 

Finally, a face detection and gaze estimation from eye pupils \cite{diego2018ddi} were showcased in the demonstration as well. 
Our planner can benefit from adding such detection technologies. Knowing the gaze allows for the inference of a targeted lane change and helps to determine if a driver is aware of critical objects. We are thus able to limit warnings to non-look cases. Since this reduces the alarm rate and workload of the driver, the social acceptance of the ADAS may be further improved.

% use section* for acknowledgment
\section*{Acknowledgment}
\noindent This work has been supported by the European Union's Hori- zon 2020 project \textit{VI-DAS}, under the grant agreement number 690772. The authors would like to thank all project partners of the \textit{VI-DAS} demonstrations.

\bibliographystyle{IEEEtran}
\bibliography{bib}

% that's all folks
\end{document}